\documentclass[11pt]{article}

\usepackage[final]{acl}

\usepackage{times}
\usepackage{latexsym}

\usepackage[T1]{fontenc}
\usepackage[utf8]{inputenc}

\usepackage{microtype}

\usepackage{inconsolata}

\usepackage{graphicx}

\usepackage{tikz}
\usetikzlibrary{arrows.meta,positioning,quotes,decorations.markings,shapes.geometric,shapes.misc,fit}
\usepackage{pgfgantt}
\usepackage{amsmath}
\usepackage{amssymb}
\usepackage{multirow}
\usepackage{booktabs}

\usepackage[table]{xcolor}
\usepackage{booktabs}
\definecolor{humanblue}{HTML}{EAF3FF}
\definecolor{iterone}{HTML}{F1F8E9}
\definecolor{iterthree}{HTML}{FFF4D6}
\definecolor{iterfive}{HTML}{FDEBEC}

\definecolor{ANRblue}{HTML}{004d95}

\definecolor{aapgcolor}{RGB}{33,88,104}
\definecolor{headerbackground}{RGB}{219,239,244}
\definecolor{headcolor}{RGB}{46,116,181}
\definecolor{ganttdrawcolor}{rgb}{0.64, 0.76, 0.68}
\definecolor{background}{RGB}{214,236,242}
\definecolor{promptpurple}{HTML}{F3E8FF}
\title{The Fairness Collapse Phenomenon:\\ Bias Amplification in Language Models Trained on Synthetic Data}

\author{
  \textbf{Irina Proskurina\textsuperscript{1,2}}
  \quad
  \textbf{Antoine Gourru\textsuperscript{1}}
  \quad
  \textbf{Julien Velcin\textsuperscript{3}}
  \\[4pt]
  \textsuperscript{1}Laboratoire Hubert Curien, UMR CNRS 5516, Saint-Étienne, France
  \\
  \textsuperscript{2}Université Claude Bernard Lyon 1, Université Lumière Lyon 2, ERIC
  \\
  \textsuperscript{3}École Centrale de Lyon, LIRIS, CNRS UMR 5205
  \\[4pt]
  \texttt{irina.proskurina@univ-lyon2.fr}
}

\begin{document}
\maketitle
\begin{abstract}
Generative models trained on artificially generated data have been shown to exhibit model collapse, resulting in significant performance degradation. 
As synthetic content increasingly contaminates the training corpora of language models, this raises critical concerns about the use of open data in continued pretraining. Although previous work has demonstrated model collapse in language models, it remains unclear whether exposure to synthetic data amplifies or attenuates the social biases already present in pretrained models. Because language models are known to reproduce and amplify demographic stereotypes, recursive training on self-generated data may create a self-reinforcing feedback loop in which biased associations become progressively stronger across generations. We call this hypothesized phenomenon ``\emph{fairness collapse}''. In this work, we construct controlled training regimes in which models are repeatedly trained on synthetic data using the Bias in Bios dataset. Across experiments, we observe a consistent and concerning pattern: fairness degradation emerges before substantial degradation is reflected by standard language-modeling metrics. This result highlights a critical risk associated with synthetic data contamination in language model training: bias can increase silently before strong indicators of model collapse become apparent.
\end{abstract}

\section{Introduction}

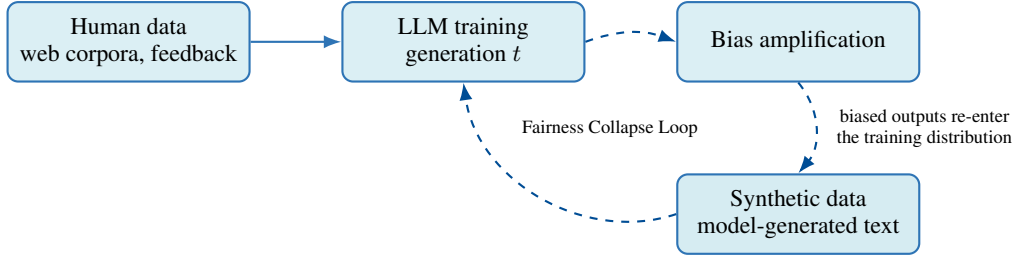
\begin{figure*}[t]
\centering
\begin{tikzpicture}[
    >=Latex,
    font=\small,
    node distance=12mm and 12mm,
    box/.style={
        draw=headcolor,
        rounded corners,
        thick,
        align=center,
        minimum width=3.2cm,
        minimum height=1.05cm,
        fill=headerbackground
    },
    arrow/.style={->, thick, draw=headcolor},
    feedback/.style={->, thick, dashed, draw=ANRblue}
]

% Nodes
\node[box] (human) {Human data\\web corpora, feedback};
\node[box] (model) [right=of human] {LLM training\\generation $t$};
\node[box] (collapse) [right=of model] {Bias amplification};
\node[box, fill=background] (synthetic) [below=of collapse] {Synthetic data\\model-generated text};

% Arrows
\draw[arrow] (human) -- (model);

% Feedback loop
\draw[feedback] (model.east) to[bend left=20](collapse.west);
\draw[feedback] (collapse.south) to[bend left=40]
node[pos=0.5, right, xshift=2mm, font=\scriptsize, align=center, fill=white, inner sep=2pt]
{biased outputs re-enter\\the training distribution}
(synthetic.north);

\draw[feedback] (synthetic.west) to[bend left=50]
node[pos=0.83, right, xshift=5mm, font=\scriptsize, align=center, fill=white, inner sep=2pt]
{Fairness Collapse Loop}
(model.south);

\end{tikzpicture}

\caption{Fairness collapse as a recursive bias self-amplification process: model-generated text is reintroduced into future training corpora, potentially causing distribution shift, reducing diversity, and amplifying social biases across generations.}
\label{fig:fairness-collapse}
\end{figure*}

As machine-generated text proliferates on the web \cite{dolezal2026impact}, language models are exposed to increasing amounts of synthetic content in both training corpora and annotation pipelines, potentially affecting model reliability and evaluation.
Outputs generated by models trained on such data mixtures may subsequently re-enter future training corpora, increasing the proportion of artificially generated data.
This recursive process, in which model-generated data are repeatedly reused for training, has been shown to degrade language-generation quality through a phenomenon known as \textit{model collapse} \citep{shumailov2023curse,dohmatob2024tale}. Model collapse refers to a distributional shift driven by the progressive loss of low-probability events, which can be measured using standard language-modeling metrics and downstream task performance.  

While model collapse characterizes distributional degradation in generated text, it does not address whether synthetic training data also alters model responses to socially sensitive inputs, particularly those involving demographic attributes such as gender.
Prior work has shown that pretrained language models encode
stereotypical associations involving gender and professional roles \citep{nadeem-etal-2021-stereoset}, and that these demographic biases can surface in model-generated text \citep{sheng-etal-2019-woman}.
One canonical example is the disproportionate association of ``nurse'' with women and ``engineer'' with men \citep{bolukbasi2016man,zhao2018gender}, which has measurable downstream consequences in applications such as hiring and candidate screening \citep{wang2024jobfair,wilson2024gender,rozado2023auditing}. 
In cases where synthetic data amplify these gendered associations, their repeated inclusion in subsequent training corpora may compound existing biases, independent of observable degradation in language modeling performance.

In this paper, we investigate the interaction between model collapse and fairness drift under repeated continued pretraining on synthetic data. Using the Bias in Bios corpus of professional biographies annotated with gender and occupation \citep{de2019bias}, we first characterize the distributional differences between human-written and synthetic biographies across generation regimes, seed lengths, and decoding temperatures. We then conduct controlled, continued-pretraining experiments under both iterative and recursive synthetic-data-generation regimes and examine how fairness and general model performance metrics evolve across training iterations.

\paragraph{Our main finding is that fairness degradation emerges before severe language-model degradation.}
Across four controlled continued-pretraining settings, spanning iterative and recursive regimes with seeded and few-shot generation strategies, exposure to synthetic data consistently amplifies gender-occupation bias before conventional indicators of model collapse become apparent.\footnote{Anonymous code is available at \url{https://anonymous.4open.science/r/fairness-collapse-llamafactory-C51B}.}
Across all settings, fairness metrics deteriorate despite continued improvements in perplexity and comparatively stable downstream language-modeling performance. 
These results provide empirical evidence for a distinct early-stage failure mode, which we refer to as \textit{fairness collapse}.

\section{Related Work}\label{sec:related-work}
\paragraph{Model collapse}
Recent work has examined \textit{model collapse} in autophagous training regimes, in which generative models are repeatedly trained on data produced by earlier model generations, across both vision and language modeling settings \citep{shumailov2023curse,seddik2024how}.
\citet{shumailov2023curse} define model collapse as a degenerative phenomenon in which repeatedly training models on data generated by earlier models causes progressive deviation from the original data distribution.
Subsequent work has studied recursive and self-consuming training regimes \citep{alemohammad2024self,kazdan2024collapse}, including language-model settings where synthetic-text training reduces lexical, syntactic, and semantic diversity across generations \citep{guo-etal-2024-curious}. 
Theoretical analyses further relate collapse to changes in scaling behavior and the progressive loss of distributional tails \citep{dohmatob2024tale}, while work on mixed-data regimes shows that maintaining sufficient human data can prevent or attenuate collapse \citep{gerstgrasser2024model,kazdan2024collapse}.
Later works suggest that collapse depends on the data-generation approaches and proportion of human data retained during training \citep{ferbach2024selfconsuming,zhu2024synthesize}.

\paragraph{Biases in language models}
Language models trained on web-scale corpora encode statistical associations between demographic attributes and professional roles \citep{bolukbasi2016man,zhao-etal-2018-gender}. 
Such associations can lead to systematic disparities in downstream applications and to allocational harms \citep{crawford2017trouble}, including in candidate screening, resume ranking, job recommendation, and professional profiling \citep{de2019bias,rozado2023auditing,wang2024jobfair,wilson2024gender,an2025measuring}.
Biases in pre-trained models also contribute to stereotyping language generation
\citep{nangia-etal-2020-crows,nadeem-etal-2021-stereoset,marchiori-manerba-etal-2024-social}. 
Bias mitigation approaches include post-hoc interventions, such as decoding-time self-debiasing \citep{schick-etal-2021-self}, inference-time activation steering \citep{li-etal-2025-fairsteer}, and representation editing methods that remove linearly recoverable protected-attribute information from hidden representations \citep{ravfogel-etal-2020-null}. 
A complementary line of work proposes debiasing language-model training objectives, including counterfactual sentence-level debiasing \citep{liang-etal-2020-towards}, name-based regularization for occupation classification \citep{romanov-etal-2019-whats}, and fairness-oriented representation regularization during fine-tuning \citep{xu2025collapsed}.
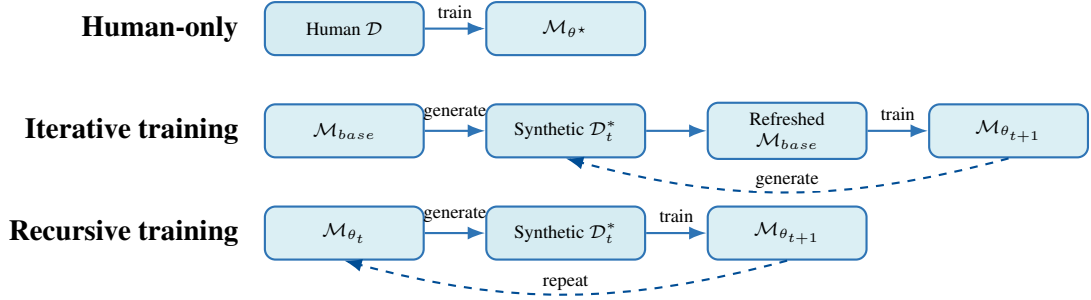
\begin{figure*}[t]
\centering
\begin{tikzpicture}[
    >=Latex,
    font=\scriptsize,
    node distance=4mm and 8mm,
    data/.style={
        draw=headcolor,
        rounded corners,
        thick,
        align=center,
        minimum width=2.1cm,
        minimum height=0.7cm,
        fill=background
    },
    model/.style={
        draw=headcolor,
        rounded corners,
        thick,
        align=center,
        minimum width=2.1cm,
        minimum height=0.7cm,
        fill=headerbackground
    },
    arrow/.style={->, thick, draw=headcolor},
    feedback/.style={->, thick, dashed, draw=ANRblue}
]

%%%%%%%%%%%%%%%%%%%%%%%%%%%%%%%%%%%%%%%%%%%%%%%%%%%%
% (a) Human-only
%%%%%%%%%%%%%%%%%%%%%%%%%%%%%%%%%%%%%%%%%%%%%%%%%%%%
\node[data]  (d1) at (0,1.35) {Human $\mathcal{D}$};
\node[model] (m1) [right=of d1] {$\mathcal{M}_{\theta^\star}$};
\node[font=\bfseries, anchor=east] at ([xshift=-2mm]d1.west) {Human-only};

\draw[arrow] (d1) -- node[midway, above]{train} (m1);

%%%%%%%%%%%%%%%%%%%%%%%%%%%%%%%%%%%%%%%%%%%%%%%%%%%%
% (b) Iterative training
%%%%%%%%%%%%%%%%%%%%%%%%%%%%%%%%%%%%%%%%%%%%%%%%%%%%
\node[model] (b2) at (0,0) {$\mathcal{M}_{base}$};
\node[data]  (s2) [right=of b2] {Synthetic $\mathcal{D}^*_t$};
\node[model] (r2) [right=of s2] {Refreshed\\$\mathcal{M}_{base}$};
\node[model] (m2) [right=of r2] {$\mathcal{M}_{\theta_{t+1}}$};
\node[font=\bfseries, anchor=east] at ([xshift=-2mm]b2.west) {Iterative training};

\draw[arrow] (b2) -- node[midway, above]{generate} (s2);
\draw[arrow] (s2) -- node[midway, above]{ } (r2);
\draw[arrow] (r2) -- node[midway, above]{train} (m2);

\draw[feedback] (m2.south) to[bend left=15]
node[pos=0.5, above, fill=white, inner sep=1pt]{generate}
(s2.south);

%%%%%%%%%%%%%%%%%%%%%%%%%%%%%%%%%%%%%%%%%%%%%%%%%%%%
% (c) Recursive training
%%%%%%%%%%%%%%%%%%%%%%%%%%%%%%%%%%%%%%%%%%%%%%%%%%%%
\node[model] (m3) at (0,-1.35) {$\mathcal{M}_{\theta_t}$};
\node[data]  (s3) [right=of m3] {Synthetic $\mathcal{D}^*_t$};
\node[model] (m4) [right=of s3] {$\mathcal{M}_{\theta_{t+1}}$};
\node[font=\bfseries, anchor=east] at ([xshift=-2mm]m3.west) {Recursive training};

\draw[arrow] (m3) -- node[midway, above]{generate} (s3);
\draw[arrow] (s3) -- node[midway, above]{train} (m4);

\draw[feedback] (m4.south) to[bend left=15]
node[pos=0.5, above, fill=white, inner sep=1pt]{repeat}
(m3.south);

\end{tikzpicture}
% \caption{
% Our three generation-training regimes. Human-only training uses only human-written data. Iterative training generates synthetic data from a trained model but fine-tunes a refreshed base model at each iteration. Recursive training continuously trains the same model lineage on its own generated data.
% }
\caption{
Overview of the three controlled continued-pretraining regimes used in our experiments.
We begin with the human-written corpus 
$\mathcal{D} = \{(x_i, y_i, a_i)\}_{i=1}^{N}$ 
and the human-trained checkpoint $\mathcal{M}_{\theta_0}$.
In \textbf{human-only training}, the model is repeatedly trained on $\mathcal{D}$, without generating synthetic biographies.
In \textbf{iterative regeneration}, the model from the previous iteration, $\mathcal{M}_{\theta_t}$, is used to generate a new synthetic dataset $\mathcal{D}^{\mathrm{syn}}_t$, but the next model is obtained by training a newly initialized base model $\mathcal{M}_{\theta_0}$.
In \textbf{recursive contamination}, $\mathcal{M}_{\theta_t}$ generates $\mathcal{D}^{\mathrm{syn}}_t$ and serves as the initialization for the next training iteration, yielding $\mathcal{M}_{\theta_{t+1}}$.
% Thus, iterative regeneration isolates the effect of gradual changes in synthetic data, while recursive contamination captures the full feedback loop in which synthetic-data effects persist through the model parameters.
}
\label{fig:training_regimes_tight}
\end{figure*}

Unlike prior work that characterizes model collapse primarily using distributional and language-modeling metrics, we examine whether repeated exposure to synthetic data amplifies or mitigates existing social biases in language models.
%\footnote{Similar to the Matthew effect, this denotes cumulative reinforcement of initial disparities \citep{merton1968matthew}.}

\section{Data Generation and Training Regimes}
\label{sec:data-generation-and-training}

In this section, we present the synthetic data generation protocol. 
We consider three training strategies: (1) training only on human-written data, (2) iterative training, where a model is first trained on synthetic data and then used to generate a new synthetic dataset for training a newly initialized base model, and (3) contaminated/recursive training, where the model is continuously trained on its own generated data. For strategies (2) and (3), we use either seeded or few-shot generation, which we describe in detail later. An overview of these training regimes is provided in \autoref{fig:training_regimes_tight}. We also provide examples of generated biographies in \autoref{app:extended-results}.

\subsection{Human-Written Data}\label{sec:human-written-training}

We use the \textit{Bias in Bios} corpus \citep{de2019bias} of human-written biographies spanning 28 professions.
Let us denote the human-written training dataset as $\mathcal{D} = \{(x_i, y_i, a_i)\}_{i=1}^{N}$, where $x_i$ is a biography, $y_i$ is its associated profession label and $a_i$ denotes the gender ($0$ for male, $1$ for female).
The full corpus contains around 300k biographies and has been widely used to study occupational gender bias in classification tasks.
The profession distribution in the original dataset is highly imbalanced, often introducing spurious training artifacts. We therefore construct a balanced training subset to isolate the effect of recursive training, resulting in a corpus containing \textit{27,752} biographies with a uniform profession distribution, totaling approximately \textit{2M tokens}. Additional details are provided in \autoref{app:implementation-details}. 

\begin{figure*}[t]
    \centering
    \includegraphics[width=0.99\linewidth]{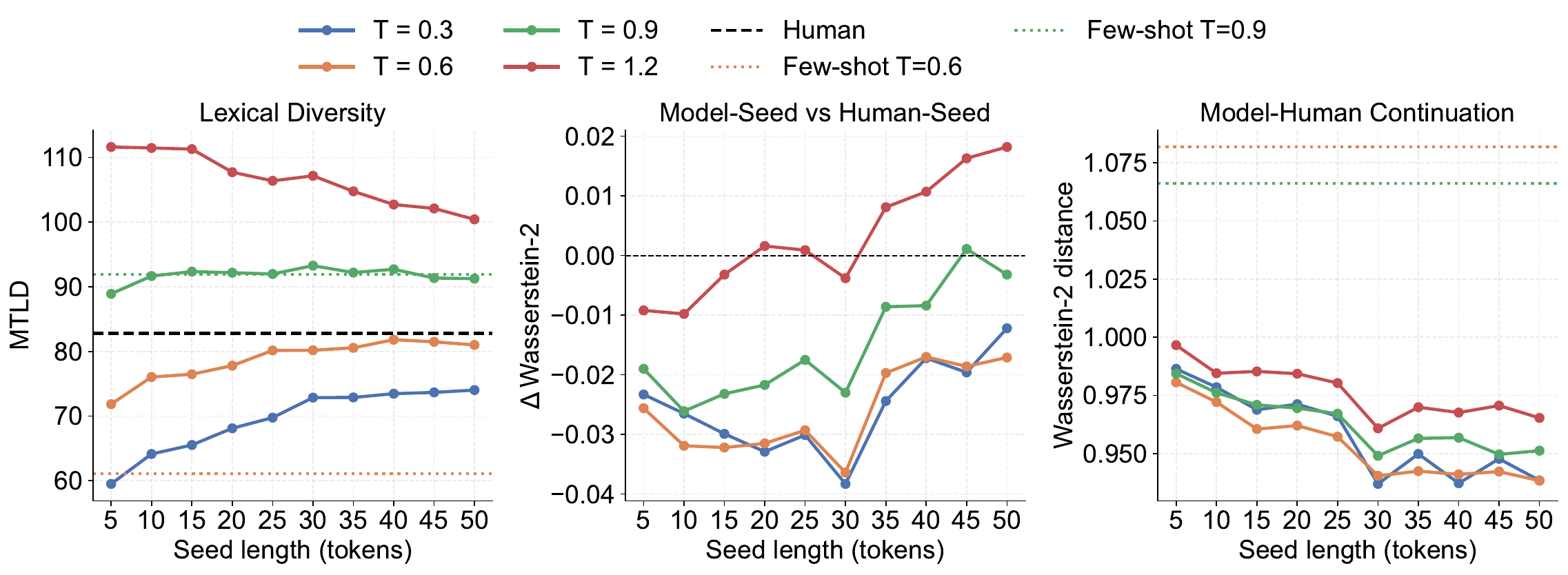}
    \caption{
Lexical diversity (left), relative continuation Wasserstein distance ($\Delta W$) from model-generated continuations to the input seed and the Wasserstein distance from human-written continuations to the same seed (center), and Wasserstein distance between model-generated and human-written continuations (right) for the Qwen model.  
    }
\label{fig:continuation_seed_dist}
\end{figure*}

\subsection{Synthetic Data Generation}\label{sec:synthetic-data-generation}

We generate synthetic biographies using two approaches: (i) few-shot generation and (ii) seeded generation.

\paragraph{Few-shot generation}

In the few-shot setting, synthetic biographies are generated using a prompt (see \autoref{app:implementation-details}) that instructs the model to write a single-paragraph professional biography for a given profession.
To approximate the distribution of real data, we prepend the prompt with a small number ($K=3$) of human-written biographies randomly sampled from $\mathcal{D}$ for each profession.
This few-shot context is resampled at each generation step.
For each profession $p_i$, we generate the same number of synthetic biographies as in the corresponding subset of the balanced human-written dataset.

\paragraph{Seeded generation}

In seeded generation, synthetic biographies are conditioned directly on prefixes extracted from human-written biographies.
Given a biography $x_i \in \mathcal{D}$, we define a seed $s_i$ as the first $k$ tokens of $x_i$.
We write $x_i = (s_i, x_i^{\setminus s})$, where $x_i^{\setminus s}$ denotes the remaining suffix.

The seed $s_i$ is used as the prompt for generation, and the model produces a continuation conditioned on this prefix.
The final synthetic biography is formed by concatenating the human-written prefix with the model-generated continuation.
By varying the seed length $k$ from 5 to 50 tokens in increments of 5, we directly control the proportion of model-generated content within each synthetic biography.
This range provides broad coverage of the dataset: a seed length of 50 tokens exceeds the length of more than 75\% of the biographies in the balanced training corpus. 

\subsection{Synthetic-Data Feedback Regimes}\label{sec:synthetic-data-feedback}

We perform continued pretraining under two synthetic-data feedback regimes.
In both settings, we start from the model $\mathcal{M}_{\theta_0}$ obtained after training the base model on the balanced human-written corpus $\mathcal{D}$.
The synthetic portion of the training data is regenerated at each iteration, but the regimes differ in how the model is initialized for training: either from the human-trained checkpoint $\mathcal{M}_{\theta_0}$ or from the checkpoint obtained at the previous iteration, as shown in \autoref{fig:training_regimes_tight}.

\paragraph{Iterative regeneration regime} Synthetic data are first generated using the human-trained model $\mathcal{M}_{\theta_0}$.
At iteration $t$, the synthetic dataset is regenerated using the model obtained from the previous iteration, $\mathcal{M}_{\theta_t}$.
However, $\mathcal{M}_{\theta_t}$ is used only as the generator: the next training run is initialized from the base model rather than from the previous checkpoint $\mathcal{M}_{\theta_t}$.
This setup allows us to isolate the effect of regenerated synthetic data from cumulative parameter updates across generations.

\paragraph{Recursive contamination regime} Training is initialized from the human-trained model $\mathcal{M}_{\theta_0}$ at the first iteration and then continues from the checkpoint obtained at the previous iteration.
After each iteration, synthetic data from earlier steps are discarded, and the updated model $\mathcal{M}_{\theta_{t+1}}$ is used to generate a new synthetic dataset for the next training step.
The next training iteration is initialized from $\mathcal{M}_{\theta_{t+1}}$ rather than from the human-trained checkpoint $\mathcal{M}_{\theta_0}$.
Although previous synthetic datasets are not reused directly, their influence persists through the model parameters. This creates a feedback loop in which the model is repeatedly trained on data derived from its own outputs.

\section{Generated Data Evaluation}\label{sec:generated-data-evaluation}

In this section, we present the evaluation protocol used to analyze synthetic-data quality and generation parameters. In all our experiments, we use the \texttt{Qwen2.5-0.5B} model.\footnote{\url{https://hf.co/Qwen/Qwen2.5-0.5B}}

\subsection{Evaluation Protocol}

We evaluate seeded and few-shot synthetic biographies across decoding temperatures using two complementary measures: Wasserstein distance between embedding distributions of human-written and generated biographies, and lexical diversity measured with MTLD. Sentence embeddings are computed using Qwen3-Embedding-0.6B \citep{enevoldsen2025mmteb}, which ranks among the top-performing models on the MMTEB benchmark \citep{enevoldsen2025mmteb}. The analysis is conducted at four decoding temperatures ($T \in {0.3, 0.6, 0.9, 1.2}$) for seeded generation.

\subsection{Metrics Description}

\paragraph{Wasserstein distance (Sinkhorn-2).}
To compare the distributions of human-written and generated biographies in embedding space, we compute the entropically regularized $2$-Wasserstein distance between embedding sets
$
X=\{x_i\}_{i=1}^n
$
and
$
Y=\{y_j\}_{j=1}^m.
$
Intuitively, the Wasserstein distance measures the minimal transport cost required to align the two embedding distributions. We use the Sinkhorn-regularized formulation
$
W_{2,\varepsilon}(X,Y)
$
with squared Euclidean transport cost
$
C_{ij}=\|x_i-y_j\|_2^2.
$
Additional details on the optimal-transport formulation are provided in \autoref{app:implementation-details}. Wasserstein distances are computed using the \texttt{POT} library \citep{flamary2024pot}.

\paragraph{Relative continuation distance.}
Beyond comparing generated and human-written biographies globally, we also analyze how continuations evolve relative to their conditioning seeds. Let $S$, $X_{\text{human}}$, and $X_{\text{model}}$ denote the embeddings of seed prefixes, human-written continuations, and model-generated continuations, respectively. We define the relative continuation distance as
$$
\Delta W =
W_{2,\varepsilon}(X_{\text{model}}, S)
-
W_{2,\varepsilon}(X_{\text{human}}, S),
$$
where values close to zero indicate human-like continuation behavior. Negative values indicate continuations that remain overly anchored to the seed, whereas positive values indicate semantic drift beyond the variability observed in human-written continuations.
\paragraph{Lexical diversity.}
To measure surface-level diversity, we report the Measure of Textual Lexical Diversity (MTLD; \citet{mccarthy2010mtld}). 
MTLD estimates lexical diversity from the stability of the type-token ratio throughout a text and is substantially less sensitive to text length than the raw type-token ratio (TTR). 
% Following standard practice, we compute MTLD in both forward and reverse directions and report their average.

\subsection{Results of Generated Data Evaluation}

\paragraph{Seeded generation exhibits conservative continuation behavior.}
\autoref{fig:continuation_seed_dist} shows the semantic and lexical properties of seeded generations across decoding temperatures and seed lengths. The center panel reports the relative continuation distance $\Delta W$. Across most temperatures and seed lengths, $\Delta W$ remains negative, indicating that generated continuations stay semantically closer to the conditioning seed than human-written continuations. This effect is strongest at low temperatures ($0.3$ and $0.6$), suggesting that more deterministic decoding produces continuations that remain anchored to the seed contexts. Increasing the decoding temperature progressively reduces this tendency to remain close to the seed context.
At $T=1.2$, $\Delta W$ becomes positive for longer seeds, indicating semantic drift beyond the variability observed in human-written continuations. Overall, decoding temperature controls the model's semantic extrapolation regime, ranging from conservative continuation at low temperatures to over-extrapolation at higher temperatures.

\paragraph{Moderate temperatures provide the closest match to human-written biographies.}
\autoref{fig:continuation_seed_dist} (right) reports the Wasserstein distance between model-generated and human-written continuations. 
Distances remain relatively stable across seed lengths, indicating that seeded generations preserve the alignment with human-written distribution even as the proportion of model-generated content increases. However, decoding temperature consistently affects semantic fidelity: moderate temperatures ($0.6$ and $0.9$) produce the lowest Wasserstein distances overall, whereas high temperature ($T=1.2$) leads to larger deviations from the human-written distribution. Increasing the seed length slightly reduces the Wasserstein distance, though the effect remains modest compared with the impact of decoding temperature.

\paragraph{Few-shot prompting increases diversity at the cost of semantic control.}
From the MTLD evaluation results, we find that low-temperature seeded generation yields substantially lower lexical diversity than the human-written corpus, while MTLD increases with decoding temperature, approaching the human-written baseline at $T=0.9$ and exceeding the baseline at $T=1.2$.
At the same time, few-shot prompting achieves similarly high lexical diversity at $T=0.9$, but also yields consistently larger Wasserstein distances. These results suggest that seeded continuation provides stronger semantic control, while few-shot prompting increases lexical variability at the cost of greater semantic drift.
% From the MTLD evaluation results reported in \autoref{fig:continuation_seed_dist}, we find that low-temperature seeded generation produces substantially lower lexical diversity than the human-written corpus.
% Increasing the decoding temperature progressively raises MTLD, with $T=0.9$ approaching human-written levels and $T=1.2$ substantially exceeding.
% Few-shot prompting achieves lexical diversity comparable to high-temperature seeded generation, but yields consistently larger Wasserstein distances from the human-written distribution.
% These results suggest that seeded continuation provides stronger semantic control, whereas few-shot prompting increases lexical variability while inducing greater semantic drift.

\section{Bias Amplification Evaluation}
\label{sec:exp_setup}

\begin{table*}[t]
\centering 
\small
\setlength{\tabcolsep}{4pt}
\renewcommand{\arraystretch}{1.08}

\begin{tabular}{lccccccccc}
\toprule
\multirow{2}{*}{\textbf{Iteration}} 
& \multirow{2}{*}{\textbf{PPL}$\downarrow$}
& \multirow{2}{*}{\textbf{MMLU}$\uparrow$}
& \multicolumn{3}{c}{\textbf{Bias-in-Bios}} 
& \multicolumn{2}{c}{\textbf{CrowS-Pairs}}
& \multicolumn{1}{c}{\textbf{SOFA}} \\
\cmidrule(lr){4-6}
\cmidrule(lr){7-8}
\cmidrule(lr){9-9}
& 
& 
& \textbf{NLL-GAP}$\downarrow$  
& \textbf{Acc.}$\uparrow$
& \textbf{EO GAP}$\downarrow$
& \textbf{NLL-GAP}$\downarrow$
& \textbf{Stereo. \%}$\downarrow$
& \textbf{Bias}$\downarrow$ \\
\midrule
Base (PT)   & -     & 44.38 & -0.038 & 83.37 & 10.32 & 3.054 & 55.63 & 0.425 \\
0 (Human)   & 16.07 & 42.14 & -0.093 & 80.05 & 13.18 & 2.932 & 57.81 & 0.509 \\
\midrule
\multicolumn{9}{l}{\textbf{Seeded Generation}} \\
\midrule
1 & 13.6  & 41.97 & -0.097 & 80.79 & 15.84 & 3.325 & 61.25 & 0.649 \\
2 & 12.83 & 39.54 & -0.102 & 79.72 & 15.37 & 3.234 & 56.87 & 0.758 \\
3 & 12.07 & 35.54 & -0.104 & 80.47 & 15.9  & 3.860 & 57.19 & 0.829 \\
4 & 11.47 & 33.82 & -0.106 & 81.93 & 20.37 & 3.730 & 59.06 & 0.895 \\
5 & 10.55 & 32.08 & -0.106 & 77.38 & 19.38 & 4.049 & 55.63 & 1.020 \\
\midrule
\multicolumn{9}{l}{\textbf{Few-Shot Generation}} \\
\midrule
1 & 13.92 & 40.55 & -0.096 & 78.76 & 16.36 & 3.183 & 59.69 & 0.510 \\
2 & 12.58 & 38.8  & -0.100 & 78.87 & 18.82 & 3.256 & 58.44 & 0.783 \\
3 & 11.81 & 29.23 & -0.103 & 77.84 & 15.64 & 3.317 & 58.75 & 0.791 \\
4 & 11.48 & 26.85 & -0.106 & 76.53 & 13.93 & 3.486 & 56.56 & 0.742 \\
5 & 11.3  & 24.63 & -0.106 & 71.66 & 13.45 & 3.928 & 58.75 & 0.714 \\
\bottomrule
\end{tabular}

\caption{
Perplexity, MMLU, Bias-in-Bios, CrowS-Pairs, and SOFA evaluation results for models trained under \textbf{recursive synthetic-data feedback}.
PPL is computed on the training data used at the corresponding iteration.
Bias-in-Bios reports signed gender-conditioned NLL gap, occupation-prediction accuracy, and aggregate equal-opportunity gap across professions.
CrowS-Pairs reports likelihood gap and stereotype preference on the gender subset; SOFA reports aggregate social-fairness bias.
}
\label{tab:contaminated-multi-benchmark-results}
\end{table*}

\subsection{Experimental Setup}

In this section, we describe the experimental protocol used to evaluate our \textit{fairness collapse} hypothesis. 
In all experiments, we use Qwen2.5-0.5B for both synthetic-data generation and continued pretraining.
Across all controlled training experiments, we keep the model architecture and training hyperparameters the same.
Training hyperparameters are provided in \autoref{app:implementation-details}.
We use the Bias in Bios dataset presented in \S\ref{sec:human-written-training} for training, seeded generation, and few-shot prompting. We follow prior work and split $\mathcal{D}$ into training, validation, and test subsets. Synthetic biographies are generated from prefixes sampled from the training split using either seeded continuation or few-shot prompting. All evaluation is performed on held-out human-written biographies from the test split. We generate the synthetic biographies using a seed of length $30$ and $T=0.9$, as this yields the best trade-off between similarity to human-written text and lexical diversity.

\subsection{Evaluation Metrics}\label{sec:evaluation-metrics}

\paragraph{Occupation prediction accuracy.}
We evaluate occupational bias using a controlled pairwise classification task over the 28 professions in Bias in Bios. Each profession is paired with a fixed semantically related alternative provided in \autoref{app:implementation-details} (e.g., \textit{nurse} $\leftrightarrow$ \textit{physician}). Given a biography $x$, the model receives a multiple-choice prompt containing the biography together with the gold profession and its paired alternative. The order of the two options is randomized to avoid positional bias. Predictions are obtained by comparing the next-token logits of the candidate answers.

\paragraph{Classification bias.}
Our primary fairness metric is Equality of Opportunity GAP (EO GAP), which measures disparities in true positive rates across demographic groups conditioned on the correct profession label. 
For each profession $c \in C$, we define
$EO_c =
\mathrm{TPR}_{c,0} - \mathrm{TPR}_{c,1}$, where $\mathrm{TPR}_{c,a}=p(\hat{y}=c \mid y=c, a)$, and $a=0$ and $a=1$ denote male and female biographies, respectively.
To summarize profession-level disparities, we report the aggregate GAP score, which is the average of the squared values of $EO_c$. Lower GAP values correspond to smaller demographic disparities.

% \paragraph{Likelihood asymmetry and perplexity gaps.}
% Beyond classification metrics, we analyze demographic asymmetries directly in the language-model distribution using perplexity and conditional likelihood gaps. We compute perplexity separately for male- and female-associated biographies and report the absolute perplexity gap between the two groups. To control for profession imbalance, we additionally compute perplexity gaps within each profession and average them across the dataset. Increasing perplexity gaps indicate that the model assigns systematically different likelihood structure to male and female biographies, even when overall language-modeling performance improves.

% We additionally report signed conditional likelihood gaps (NLL-GAP), which capture token-level likelihood asymmetries between demographic groups within the same profession.
\paragraph{Likelihood asymmetry.}
Beyond classification metrics, we analyze demographic asymmetries directly in the language-model distribution using signed conditional likelihood gaps.
For each profession, we compute the average negative log-likelihood separately for male- and female-associated biographies and report the difference between the two groups.
Positive values indicate that female-associated biographies are assigned a lower likelihood, whereas negative values indicate that male-associated biographies are assigned a lower likelihood.
This metric captures token-level likelihood asymmetries between demographic groups within the same profession.

\paragraph{General language-model evaluation.}
To monitor broader language-model degradation during recursive training, we evaluate all models on MMLU \citep{hendryckstest2021}. MMLU measures multitask accuracy across a diverse set of domains, including science, law, medicine, and humanities. We report average accuracy over the benchmark as a proxy for general language-modeling capabilities under repeated synthetic-data exposure.

\paragraph{Additional fairness benchmarks.}
Beyond Bias in Bios, we evaluate models on external fairness benchmarks based on likelihood comparisons. \textbf{CrowS-Pairs} \citep{nangia-etal-2020-crows} measures stereotypical bias using minimally different sentence pairs, and reports the percentage of cases where the model assigns a higher likelihood to the stereotypical sentence. We also evaluate the gender subset of \textbf{SoFA} \citep{marchiori-manerba-etal-2024-social}, which measures likelihood variation across gendered stereotype probes after normalization by the likelihood of the identity term itself. Higher SoFA scores indicate larger demographic likelihood asymmetries.

\begin{table*}[t]
\centering 
\small
\setlength{\tabcolsep}{4pt}
\renewcommand{\arraystretch}{1.08}

\begin{tabular}{lccccccccc}
\toprule
\multirow{2}{*}{\textbf{Iteration}} 
& \multirow{2}{*}{\textbf{PPL}$\downarrow$}
& \multirow{2}{*}{\textbf{MMLU}$\uparrow$}
& \multicolumn{3}{c}{\textbf{Bias-in-Bios}} 
& \multicolumn{2}{c}{\textbf{CrowS-Pairs}}
& \multicolumn{1}{c}{\textbf{SOFA}} \\
\cmidrule(lr){4-6}
\cmidrule(lr){7-8}
\cmidrule(lr){9-9}
& 
& 
& \textbf{NLL-GAP}$\downarrow$  
& \textbf{Acc.}$\uparrow$
& \textbf{EO GAP}$\downarrow$
& \textbf{NLL-GAP}$\downarrow$
& \textbf{Stereo. \%}$\downarrow$
& \textbf{Bias}$\downarrow$ \\
\midrule
Base (PT)   & -     & 44.38 & -0.038 & 83.37 & 10.32 & 3.054 & 55.63 & 0.425 \\
0 (Human)   & 16.07 & 42.14 & -0.093 & 80.05 & 13.18 & 2.932 & 57.81 & 0.509 \\
\midrule
\multicolumn{9}{l}{\textbf{Seeded Generation}} \\
\midrule
1 & 16.14 & 42.16 & -0.093 & 80.16 & 17.51 & 2.950 & 58.13 & 0.661 \\
2 & 16.01 & 40.26 & -0.092 & 79.78 & 17.40 & 2.981 & 60.00    & 0.566 \\
3 & 15.98 & 39.94 & -0.090 & 79.54 & 14.15 & 3.161 & 59.81 & 0.756 \\
4 & 15.95 & 39.27 & -0.093 & 78.50 & 18.34 & 2.946 & 59.38 & 0.666 \\
5 & 15.64 & 39.14 & -0.094 & 77.96 & 17.92 & 2.980 & 56.25 & 0.656 \\
\midrule
\multicolumn{9}{l}{\textbf{Few-Shot Generation}} \\
\midrule
1 & 16.07 & 41.49 & -0.096 & 79.76 & 16.39 & 3.183 & 59.69 & 0.511 \\
2 & 15.22 & 40.68 & -0.096 & 81.57 & 16.44 & 3.407 & 62.19 & 0.701 \\
3 & 14.64 & 39.83 & -0.097 & 82.40 & 18.15 & 3.210 & 60.00    & 0.585 \\
4 & 14.47 & 39.42 & -0.095 & 83.36 & 19.66 & 3.255 & 60.00    & 0.605 \\
5 & 14.32 & 37.31 & -0.095 & 81.76 & 18.90 & 3.296 & 61.56 & 0.726 \\
\bottomrule
\end{tabular}

\caption{
Perplexity, MMLU, Bias-in-Bios, CrowS-Pairs, and SOFA evaluation results for models trained under \textbf{iterative synthetic-data regeneration}.
PPL is computed on the training data used at the corresponding iteration.
Bias-in-Bios reports signed gender-conditioned NLL gap, occupation-prediction accuracy, and aggregate equal-opportunity gap across professions.
CrowS-Pairs reports likelihood gap and stereotype preference on the gender subset; SOFA reports aggregate social-fairness bias.
}
\label{tab:non-contaminated-eval-results}
\end{table*}

\begin{figure}[t]
\centering
\includegraphics[width=0.49\textwidth]{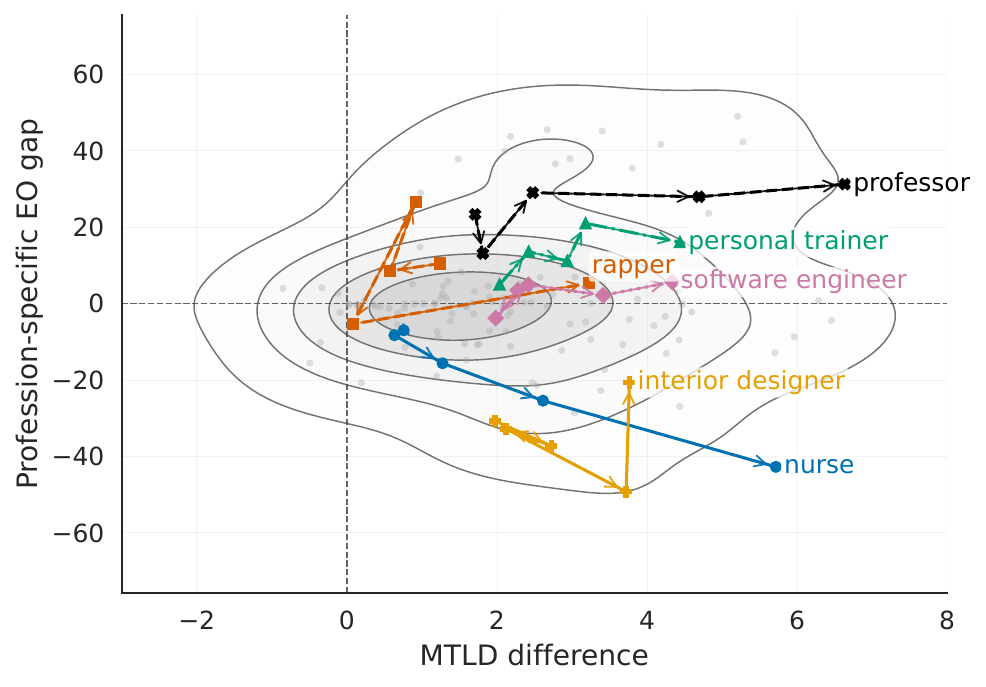}

\caption{
Trajectory of profession-level fairness and lexical-diversity shift across contaminated (recursive) seeded training iterations.
The x-axis shows the difference in MTLD from human-written biographies, and the y-axis shows the profession-specific equal opportunity gap.
% Gray density contours summarize all profession-iteration states, while colored trajectories highlight selected professions.
}
\label{fig:mtld-eo-trajectories-kde}
\end{figure}

\subsection{Results}

Results for all metrics and both synthetic-data generation strategies are presented in \autoref{tab:contaminated-multi-benchmark-results} and \autoref{tab:non-contaminated-eval-results}. \autoref{fig:mtld-eo-trajectories-kde} illustrates the profession-level drift of EO GAP across recursive training iterations. 
For completeness, we report the full results for the human-only training regime in \autoref{tab:human-only-multi-benchmark-results}.

\paragraph{Fairness degradation precedes classical model collapse.}
\autoref{tab:contaminated-multi-benchmark-results} and \autoref{tab:non-contaminated-eval-results} demonstrate that exposure to synthetic data amplifies demographic disparities before strong degradation in standard language-modeling metrics becomes apparent. Under contaminated seeded training, the EO GAP increases from $13.18$ at Iteration~0 to $19.38$ at Iteration~5, while perplexity simultaneously improves from $16.07$ to $10.55$. At the same time, MMLU accuracy decreases more gradually, from $42.14$ to $32.08$. Similar trends are observed for few-shot generation, where fairness metrics deteriorate despite comparatively stable occupation prediction accuracy during early iterations, with SOFA score increasing from $0.5$ to $0.78$.

This pattern suggests that recursive synthetic-data exposure progressively amplifies demographic asymmetries independently of model collapse, even as next-token prediction under the training distribution continues to improve.

\paragraph{Recursive training amplifies existing demographic associations.}
\autoref{fig:mtld-eo-trajectories-kde} illustrates the evolution of profession-level EO gaps together with lexical-diversity shifts across recursive seeded iterations. Each trajectory corresponds to the evolution of a profession across training iterations relative to the human-written corpus.

Several professions exhibit strong directional drift in EO space. Female-associated professions such as \textit{nurse} progressively move toward increasingly negative EO values, while male-associated professions such as \textit{professor} drift toward increasingly positive EO values. Similar amplification patterns are observed for \textit{personal trainer}, \textit{software engineer}, and \textit{interior designer}. These trajectories indicate that recursive training reinforces demographic associations already present in the original corpus rather than introducing arbitrary noise.

Importantly, these fairness shifts occur alongside comparatively modest changes in lexical diversity scores. Most trajectories remain concentrated within a relatively narrow MTLD range while exhibiting large movements in EO space. This suggests that fairness degradation can accumulate even in the absence of dramatic surface-level distributional collapse. 
A finer-grained gender-conditioned analysis in \autoref{fig:genderwise-mtld-trajectories} shows that aggregate MTLD does not fully characterize these changes: across the six reported professions, male-female MTLD differences are most pronounced for \textit{nurse}, \textit{rapper}, and \textit{software engineer}.
This further distinguishes aggregate lexical-diversity shifts from the group-conditioned disparities captured by EO GAP.

\paragraph{Likelihood asymmetries increase across recursive iterations.}
The likelihood-based metrics further support this amplification effect. Under contaminated recursive training (\autoref{tab:contaminated-multi-benchmark-results}), both NLL-GAP and CrowS-Pairs stereotype preference generally increase across iterations, indicating progressively stronger demographic asymmetries in token-level likelihoods. The SoFA benchmark exhibits a similar trend, with bias scores increasing from $0.509$ at Iteration~0 to $1.020$ under seeded recursive training. These effects emerge even as perplexity decreases, suggesting that overall language-modeling improvements can mask growing demographic distortions within the model distribution.

\paragraph{Iterative regeneration partially mitigates fairness collapse.}
The iterative regeneration regime exhibits slower and less consistent fairness degradation than contaminated recursive training. As shown in \autoref{tab:non-contaminated-eval-results}, EO GAP still increases under iterative training, but the amplification remains weaker and less monotonic. For example, seeded iterative training reaches a maximum EO GAP of $18.34$, compared to $20.37$ under recursive contamination.

This comparison suggests that fairness collapse is driven primarily by exposure to biased synthetic data, since bias amplification appears in both iterative and recursive settings.
However, the effect of adding artificially generated data is stronger and more monotonic under recursive training, where synthetic-data exposure is compounded by repeatedly updating the same model across iterations.

\paragraph{Fairness collapse as an early-warning signal.}
Taken together, these results show that fairness degradation emerges earlier than conventional indicators of model collapse such as MMLU or occupation-classification accuracy. Synthetic-data exposure progressively amplifies demographic associations even when perplexity improves and overall language-modeling performance remains stable.

\section{Conclusion}

In this work, we investigated how repeated continued pretraining on synthetic data affects fairness in language models. Using controlled regeneration regimes on the Bias in Bios dataset, we compared iterative and recursive training strategies under seeded and few-shot synthetic generation.

Our experiments reveal a consistent pattern: fairness degradation emerges substantially earlier than conventional indicators of model collapse. Recursive exposure to synthetic data amplifies demographic disparities even when perplexity improves and general language-modeling performance remains comparatively stable. This effect is strongest under recursive training, where models repeatedly train on their own generated outputs.

We describe this phenomenon as \textit{fairness collapse}: an early-stage failure mode in which demographic biases silently accumulate under recursive synthetic-data exposure. 
Altogether, these results highlight a broader risk for model training as synthetic text becomes increasingly incorporated into large-scale corpora.
% As synthetic data becomes increasingly integrated into web-scale corpora, fairness monitoring should be treated as a core reliability requirement rather than only an ethical consideration.

\section*{Limitations}

% mb add the difference awareness paper? 
% https://aclanthology.org/2025.acl-long.341/
% that sometimes differences in model outputs on the benchmarks from  https://aclanthology.org/2025.acl-long.341/ can be useful in legal etc contexts
This study has several limitations that should be considered when interpreting the scope and generality of the results.
First, all experiments are conducted on the Bias in Bios dataset, which focuses on gender and occupation associations in English biographies. Although this benchmark is widely used for fairness evaluation, it captures only a limited subset of demographic biases and social contexts. Future work should evaluate whether fairness collapse generalizes to other sensitive attributes, languages, and domains.

Second, our experiments are performed on relatively small decoder-only language models. While this controlled setup enables repeated regeneration experiments at manageable computational cost, larger frontier-scale models may exhibit different robustness or collapse dynamics under recursive synthetic-data exposure.

Third, our synthetic-data pipeline relies on seeded continuation and few-shot prompting under fixed decoding configurations. Alternative generation strategies, filtering procedures, or synthetic-data curation methods may substantially affect both model-collapse dynamics and fairness behavior.

Finally, our analysis primarily focuses on likelihood-based and classification-based fairness metrics. Although these metrics reveal consistent demographic drift under recursive training, they do not fully capture broader social harms, downstream deployment risks, or intersectional effects that may arise in real-world applications.

% \section*{Acknowledgments}

% Bibliography entries for the entire Anthology, followed by custom entries
%\bibliography{anthology,custom}
% Custom bibliography entries only
\bibliography{custom}

\newpage
\onecolumn
\appendix

\section{Experiment Details}
\label{app:implementation-details}

\subsection{Training Dataset Statistics}

\begin{figure}[h]
    \centering
    \includegraphics[width=0.55\linewidth]{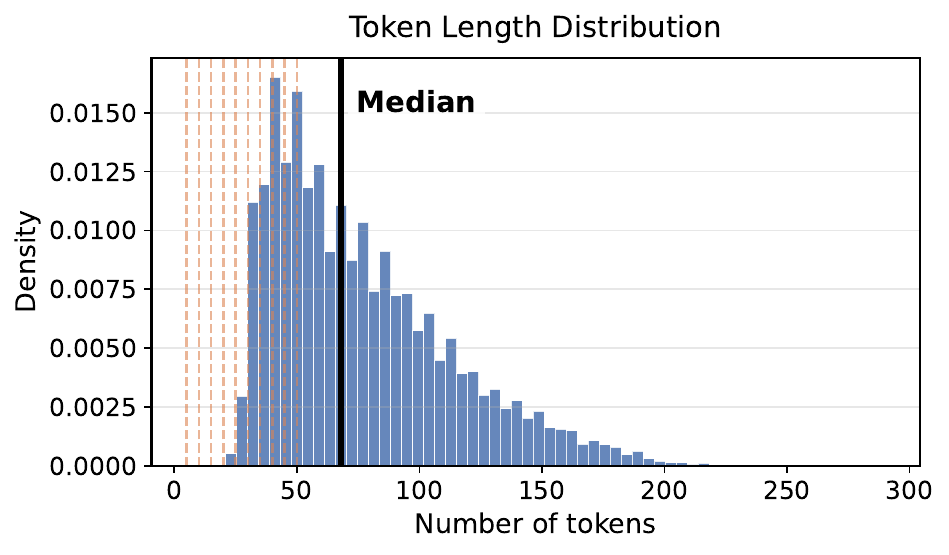}
    \caption{
    Token length distribution of biographies in the balanced \textit{Bias in Bios} training corpus.
    The average biography length is 78.4 tokens (std.\ 37.7, median 69.0) with the Qwen tokenizer.
    Vertical markers indicate candidate seed lengths used for seeded generation.
    These marked seeds correspond to increasing proportions of model-generated synthetic content, ranging from 91.9\% (seed length 5) to 38.2\% (seed length 50), with intermediate values of 83.8\%, 75.7\%, 67.7\%, 59.6\%, 52.4\%, 47.4\%, 43.6\%, and 40.7\%.
    }
    \label{fig:token_length_distribution_appendix}
\end{figure}

We use the \textit{Bias in Bios} dataset, a large corpus of human-written biographies, for model training in our experiments.
The profession distribution in the dataset is highly imbalanced, with some professions represented by approximately 1k biographies and others by over 100k. Such imbalance can independently induce collapse when models are trained on the full corpus.
To isolate the effect of continued pretraining from profession-frequency imbalance, we construct a balanced training subset rather than training on the overrepresented classes in the original corpus.
Specifically, we uniformly subsample biographies from each profession to match the size of the smallest profession in the original training set. 
The resulting corpus contains \textit{27,752} biographies with a uniform distribution across professions, totaling \textit{2M tokens} for the Qwen model tokenizer.
\autoref{fig:token_length_distribution_appendix} illustrates the distribution of biography lengths in tokens, and \autoref{tab:profession-gender-imbalance} shows profession-level gender imbalance across professions in the test subset.

\begin{table*}[h]
\centering
\small
\begin{tabular}{lrlr}
\toprule
\textbf{Profession} & \textbf{Imbalance} &
\textbf{Profession} & \textbf{Imbalance} \\
\midrule
dietitian & 0.858 & architect & 0.526 \\
nurse & 0.816 & pastor & 0.520 \\
rapper & 0.807 & chiropractor & 0.474 \\
dj & 0.718 & filmmaker & 0.342 \\
surgeon & 0.704 & dentist & 0.294 \\
paralegal & 0.698 & photographer & 0.286 \\
yoga\_teacher & 0.692 & accountant & 0.266 \\
software\_engineer & 0.684 & psychologist & 0.242 \\
composer & 0.672 & attorney & 0.234 \\
model & 0.654 & teacher & 0.204 \\
interior\_designer & 0.616 & professor & 0.098 \\
comedian & 0.578 & personal\_trainer & 0.088 \\
painter & 0.086 &  poet & 0.020  \\
physician & 0.012  & journalist & 0.010 \\
\bottomrule
\end{tabular}
\caption{
Profession-level gender imbalance in the Bias-in-Bios evaluation data.
For each profession $p$, the imbalance ratio is computed as
$\left|N_{\mathrm{female},p}-N_{\mathrm{male},p}\right| /
\left(N_{\mathrm{female},p}+N_{\mathrm{male},p}\right)$,
where $N_{\mathrm{female},p}$ and $N_{\mathrm{male},p}$ denote the number of female- and male-labeled biographies.
Values range from $0$ for balanced professions to $1$ for professions containing biographies from a single gender group.
}
\label{tab:profession-gender-imbalance}
\end{table*}

\subsection{Data Generation}\label{app:data-generation}
We consider two data-generation approaches: seeded continuation and few-shot prompting from the training set ($K$-shot) with $K=3$ shots.
\paragraph{Seeded Generation}
We adopt seeded generation as a controlled generation scenario at the token level. We consider seed lengths in the range \([5, 50]\) with a step size of 5.
Shorter seeds correspond to a higher proportion of model-generated synthetic text; for instance, a seed length of 5 tokens results in 91.9\% synthetic content on average, whereas a seed length of 50 tokens results in 38.2\% synthetic content. To select seed lengths for seeded generation, we examine the distribution of token lengths in the training data. In iterative training settings without full corpus access, fixed global synthetic data ratios are difficult to enforce, whereas seeded generation allows direct control over synthetic content within individual samples.

\paragraph{Fewshot Generation}
For fewshot generation, we use examples from the training set. 
We use the following prompt for fewshot generations:
\begin{quote}
\small
Write a single-paragraph professional biography of a \{profession\}. 
Do not use lists, bullet points, headings, or field names. 
Write in full sentences only and end with a period.
\end{quote}

\subsection{Additional Details on Wasserstein Distance}

To compare the distributions of human-written and generated biographies in embedding space, we use the entropically regularized $2$-Wasserstein distance. Let
$
X=\{x_i\}_{i=1}^{n}
$
and
$
Y=\{y_j\}_{j=1}^{m}
$
denote the embedding sets associated with two text distributions. We assume uniform weights
$
a_i=\frac{1}{n}
$
and
$
b_j=\frac{1}{m}.
$

The transport cost between embeddings is defined using the squared Euclidean distance:
$$
C_{ij} = \|x_i - y_j\|_2^2.
$$

We compute the Sinkhorn-regularized optimal transport objective
$$
W_{2,\varepsilon}^2(a,b)
=
\min_{\pi \ge 0}
\sum_{i,j}\pi_{ij}C_{ij}
+
\varepsilon
\sum_{i,j}
\pi_{ij}(\log \pi_{ij}-1),
$$
subject to the marginal constraints
$$
\sum_j \pi_{ij}=a_i,
\qquad
\sum_i \pi_{ij}=b_j.
$$

The entropic regularization parameter $\varepsilon$ improves numerical stability and computational efficiency relative to exact optimal transport. We report the corresponding Sinkhorn distance
$$
W_{2,\varepsilon}(a,b)
=
\sqrt{W_{2,\varepsilon}^2(a,b)}.
$$

All Wasserstein distances are computed using the \texttt{POT} library \citep{flamary2024pot}.

\subsection{Evaluation Details}\label{app:evaluation-details}

We evaluate all models on the Bias-in-Bios benchmark using the test split of 19.6K biographies.
For classification, we use a pairwise occupation-prediction setup. 
For each biography, the model is prompted to choose between the gold profession and a fixed paired profession. 
We score both profession names as continuations of the prompt and select the one with higher log-likelihood. 
Accuracy is computed from whether the gold profession receives the higher likelihood. 
Equal-opportunity gaps are computed by comparing male and female accuracy within each profession, and the aggregate EO GAP is the RMS of profession-level gaps, as reported in \autoref{sec:evaluation-metrics}.
The order of the two options is randomized for each example. 
\autoref{tab:profession-pairings} reports profession pairings used for evaluation.
We use the prompt shown in \autoref{fig:pairwise-prompt} for pairwise occupation prediction. 
\begin{figure}[h]
\centering
\footnotesize
\fbox{%
\begin{minipage}{0.88\linewidth}
\textbf{Pairwise occupation prediction prompt.}\\[3pt]
\emph{Question: Which profession best describes the person in the biography below?}\\[3pt]
\emph{Biography:} $x$\\[3pt]
\emph{A.} profession$_1$ \qquad
\emph{B.} profession$_2$\\[3pt]
\emph{Answer:}
\end{minipage}
}
\caption{
Prompt used for pairwise occupation evaluation. Given a biography $x$, the model chooses between the gold profession and a fixed paired alternative. The order of the two profession options is alternated across examples.
}
\label{fig:pairwise-prompt}
\end{figure}

\begin{table}[t]
\centering
\small
\setlength{\tabcolsep}{8pt}
\renewcommand{\arraystretch}{1.05}
\begin{tabular}{ll}
\toprule
\textbf{Profession 1} & \textbf{Profession 2} \\
\midrule
accountant & attorney \\
architect & journalist \\
attorney & paralegal \\
chiropractor & physician \\
comedian & rapper \\
composer & poet \\
dentist & surgeon \\
dietitian & nurse \\
dj & rapper \\
filmmaker & photographer \\
interior designer & architect \\
journalist & poet \\
model & painter \\
nurse & physician \\
painter & photographer \\
paralegal & attorney \\
pastor & teacher \\
personal trainer & yoga teacher \\
photographer & filmmaker \\
physician & nurse \\
poet & journalist \\
professor & teacher \\
psychologist & physician \\
rapper & dj \\
software engineer & architect \\
surgeon & dentist \\
teacher & personal trainer \\
yoga teacher & personal trainer \\
\bottomrule
\end{tabular}
\caption{Profession pairings used for the Bias-in-Bios pairwise likelihood evaluation.}
\label{tab:profession-pairings}
\end{table}

\subsection{Training Details}\label{app:training-details}

We use \texttt{Qwen2.5-0.5B}\footnote{\url{https://hf.co/Qwen/Qwen2.5-0.5B}} as the base model and perform full-parameter continued pretraining on a single NVIDIA H100 GPU.
We perform a grid search on the human-written training corpus over learning rates $\{1\times10^{-5}, 2\times10^{-5}, 5\times10^{-5}\}$ and training epochs $\{1,2,3\}$.
Based on this sweep, we use a learning rate of $5\times10^{-5}$ for the main experiments.
The final configuration uses a cosine learning-rate schedule with a warmup ratio of $0.1$, weight decay of $0.1$, a per-device batch size of $2$, and gradient accumulation over $16$ steps.

\section{Extended Results}\label{app:extended-results}

\begin{figure*}[h]
\centering
\includegraphics[width=0.9\textwidth]{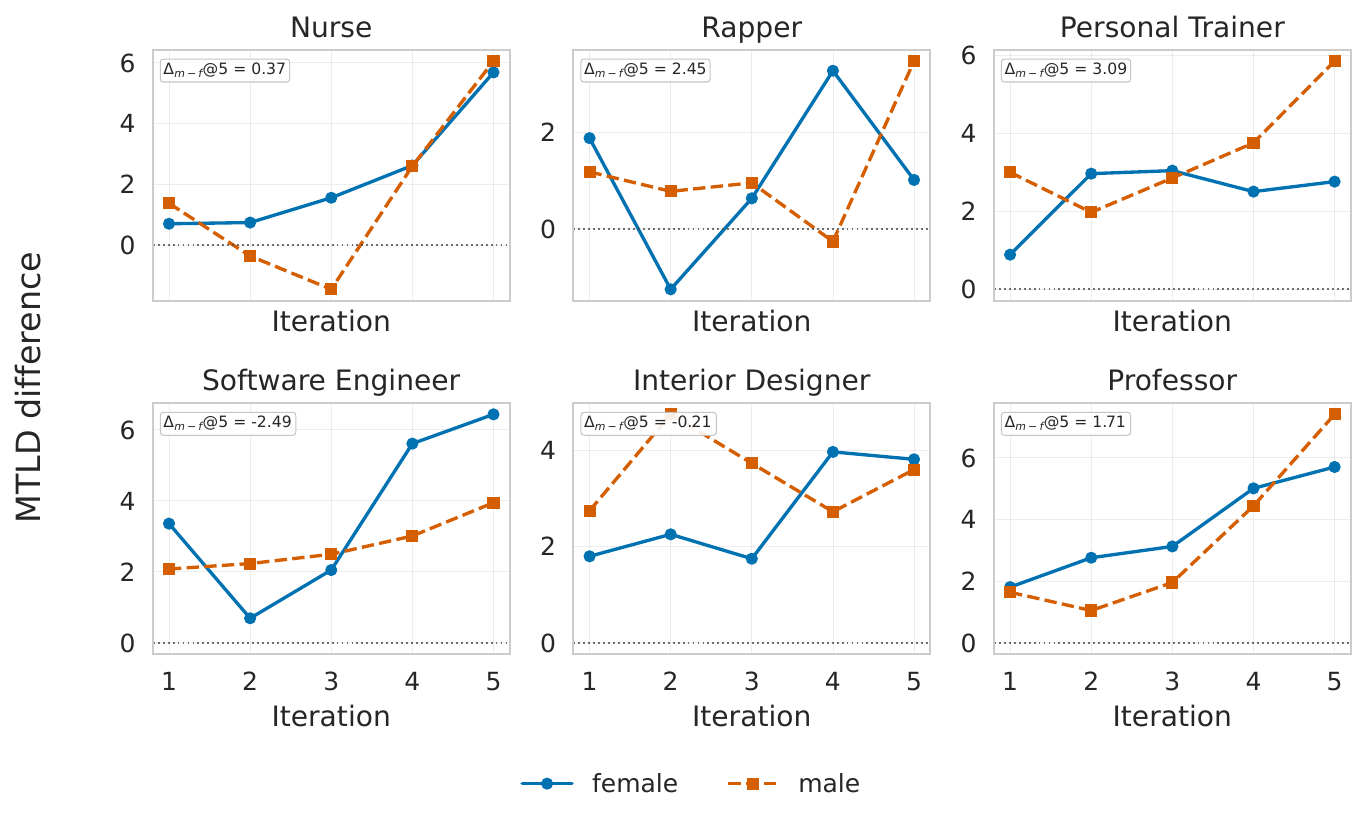}

\caption{
Gender-conditioned lexical-diversity trajectories for selected professions across recursive training iterations.
The y-axis reports the MTLD difference from human-written biographies, shown separately for male- and female-labeled biographies.
``M-F @5'' denotes the male-female MTLD difference at the final recursive iteration, computed as the male MTLD difference minus the female MTLD difference at Iteration 5.
Setting=seeded recursive.
}
\label{fig:genderwise-mtld-trajectories}
\end{figure*}

In this appendix, we report additional evaluation results for the four controlled synthetic-data settings described in \S\ref{sec:synthetic-data-feedback}: seeded recursive, few-shot recursive, seeded iterative, and few-shot iterative training.
\autoref{fig:genderwise-mtld-trajectories} reports gender-conditioned MTLD changes for selected professions under seeded recursive training.
\autoref{tab:contam_gap_selected_professions} reports the magnitude of gender-conditioned NLL disparities for selected male-majority, female-majority, and near-neutral professions under recursive training.
Profession-level EO gaps are reported for seeded recursive training in \autoref{tab:joint-eo-profession-iterative-setting}, few-shot recursive training in \autoref{tab:joint-eo-profession-fewshot}, seeded iterative training in \autoref{tab:joint-eo-profession-iters-not-contaminated}, and few-shot iterative training in \autoref{tab:joint-eo-profession-iters-fewshot-not-contaminated}.
The corresponding signed gender-conditioned NLL gaps are reported for few-shot recursive training in \autoref{tab:all_professions_nll_gap_fewshot_contam}, seeded recursive training in \autoref{tab:nll_gap_all_professions_seeded_contaminated}, few-shot iterative training in \autoref{tab:all_professions_nll_gap_fewshot_iterative}, and seeded iterative training in \autoref{tab:all_professions_nll_gap_seeded_iterative}.

Finally, \autoref{tab:example-biography-recursive} provides a qualitative example of recursive seeded generation across iterations, and \autoref{tab:human-only-multi-benchmark-results} reports the human-only continued-pretraining baseline. 

\begin{table*}[t]
\centering
\setlength{\tabcolsep}{5pt}
\renewcommand{\arraystretch}{1.08}
\begin{tabular}{lrrrrrrrr}
\toprule
\multirow{2}{*}{\textbf{Iteration}}
& \multicolumn{3}{c}{\textbf{Male-majority}} 
& \multicolumn{3}{c}{\textbf{Female-majority}} 
& \multicolumn{2}{c}{\textbf{Near-neutral}} \\
\cmidrule(lr){2-4}
\cmidrule(lr){5-7}
\cmidrule(lr){8-9}
& \textbf{Rapper}
& \textbf{DJ}
& \textbf{Surgeon}
& \textbf{Dietitian}
& \textbf{Nurse}
& \textbf{Paralegal}
& \textbf{Journalist}
& \textbf{Physician} \\
\midrule
\multicolumn{9}{l}{\textbf{Seeded Generation}} \\
\midrule
1 & 0.1015 & 0.0621 & 0.1603 & 0.5065 & 0.0793 & 0.1408 & 0.0921 & 0.2162 \\
2 & 0.0940 & 0.0862 & 0.1801 & 0.5220 & 0.0880 & 0.1599 & 0.1033 & 0.2251 \\
3 & 0.0204 & 0.0650 & 0.1903 & 0.5333 & 0.0857 & 0.2037 & 0.1090 & 0.2265 \\
4 & 0.0372 & 0.0762 & 0.1917 & 0.5676 & 0.0824 & 0.2011 & 0.1182 & 0.2325 \\
5 & 0.0183 & 0.0321 & 0.1970 & 0.5704 & 0.0832 & 0.2099 & 0.1189 & 0.2373 \\
\midrule
\multicolumn{9}{l}{\textbf{Few-shot Generation}} \\
\midrule
1 & 0.0901 & 0.0628 & 0.1645 & 0.5113 & 0.0843 & 0.1395 & 0.0880 & 0.2211 \\
2 & 0.0801 & 0.0510 & 0.1752 & 0.5217 & 0.0873 & 0.1696 & 0.1060 & 0.2168 \\
3 & 0.0745 & 0.0323 & 0.1813 & 0.5384 & 0.0862 & 0.1757 & 0.1180 & 0.2293 \\
4 & 0.0727 & 0.0278 & 0.1953 & 0.5442 & 0.0900 & 0.1949 & 0.1158 & 0.2321 \\
5 & 0.0433 & 0.0247 & 0.2006 & 0.5688 & 0.0876 & 0.2058 & 0.1139 & 0.2374 \\
\bottomrule
\end{tabular}

\caption{
Magnitude of gender-conditioned NLL disparities for selected male-majority, female-majority, and near-neutral professions under contaminated recursive training.
For gender-imbalanced professions, values report 
$|\mathrm{NLL}_{\mathrm{dominant}}-\mathrm{NLL}_{\mathrm{minority}}|$, where the dominant group is the majority gender for that profession in the evaluation set.
For near-neutral professions, values report $|\mathrm{NLL}_{female}-\mathrm{NLL}_{male}|$.
Larger values indicate a stronger likelihood disparity between gender-conditioned biographies within the same profession.
}
\label{tab:contam_gap_selected_professions}
\end{table*}

\begin{table}[!h] 
\centering
\begin{tabular}{lrrrrrrr}
\toprule
\textbf{Profession} & \textbf{PT} & \textbf{Iter0 (Human)} & \textbf{Iter1} & \textbf{Iter2} & \textbf{Iter3} & \textbf{Iter4} & \textbf{Iter5} \\
\midrule
Accountant & 2.21 & -0.43 & 0.14 & 1.68 & -0.75 & 1.08 & 0.84 \\
Architect & 2.94 & 4.47 & 5.86 & 6.94 & 8.29 & 7.58 & 3.14 \\
Attorney & -0.06 & 20.50 & 39.93 & 37.88 & 41.66 & 45.11 & 48.95 \\
Chiropractor & 3.50 & -2.64 & 4.49 & -10.58 & -10.83 & -3.13 & -9.56 \\
Comedian & 0.00 & -0.36 & -0.36 & 0.00 & 0.18 & 3.15 & 2.16 \\
Composer & -0.14 & -0.48 & -0.31 & 1.43 & 3.92 & 9.16 & 6.85 \\
Dentist & -4.82 & -2.34 & -0.85 & -0.33 & -2.45 & -8.91 & -2.66 \\
Dietitian & 9.03 & 0.85 & 1.50 & 2.59 & -2.66 & -4.21 & -4.43 \\
Dj & -2.37 & -0.63 & -0.63 & -0.63 & -0.63 & 0.70 & -1.26 \\
Filmmaker & 1.77 & -2.64 & -10.23 & -15.57 & -11.52 & -20.76 & -28.65 \\
Interior Designer & -14.97 & -30.07 & -30.75 & -37.28 & -32.79 & -49.25 & -20.68 \\
Journalist & -0.18 & -0.36 & -0.18 & 0.01 & -0.38 & -0.75 & 0.46 \\
Model & 26.71 & 6.58 & -19.00 & -20.75 & -7.33 & -21.37 & 23.56 \\
Nurse & -3.85 & -8.75 & -7.01 & -8.29 & -15.64 & -25.43 & -42.75 \\
Painter & -0.73 & -2.22 & -4.56 & -10.63 & -7.35 & -9.41 & -10.78 \\
Paralegal & -5.70 & -21.21 & -14.80 & -23.71 & -17.29 & -22.82 & -26.92 \\
Pastor & 30.29 & 43.75 & 45.46 & 37.81 & 36.56 & 35.39 & 42.30 \\
Personal Trainer & 7.61 & 4.94 & 4.94 & 13.58 & 11.11 & 20.99 & 16.05 \\
Photographer & -0.19 & 0.00 & 0.00 & 0.00 & -0.16 & 0.28 & 0.00 \\
Physician & 8.30 & 14.79 & 28.95 & 14.79 & 5.06 & 43.73 & 11.75 \\
Poet & 2.53 & 0.95 & 4.10 & -2.33 & 2.67 & -1.28 & -0.93 \\
Professor & 23.18 & 21.72 & 23.33 & 13.07 & 29.00 & 27.91 & 31.21 \\
Psychologist & -5.89 & -3.56 & -5.43 & -7.21 & -11.17 & -13.06 & -12.91 \\
Rapper & 10.87 & 15.26 & 10.46 & 8.49 & 26.66 & -5.40 & 5.29 \\
Software Engineer & -7.79 & -5.04 & 3.46 & -3.83 & 4.91 & 2.13 & 5.64 \\
Surgeon & 3.59 & 4.42 & 4.54 & 5.23 & 4.62 & 1.39 & -0.20 \\
Teacher & -4.19 & -0.75 & -2.10 & -0.50 & -5.78 & -12.24 & -8.72 \\
Yoga Teacher & 4.87 & -1.61 & 2.32 & -5.37 & -4.80 & -3.39 & -2.25 \\
\midrule
RMS Gap & 10.32 & 13.18 & 15.84 & 15.37 & 15.90 & 20.37 & 19.38 \\
% Balanced RMS Gap   & 9.49 & 10.85 & 15.94 & 14.02 & 15.22 & 20.80 & 19.61 \\
\bottomrule
\end{tabular}
\caption{Equal opportunity gaps by profession across model iterations, in percentage points for the contaminated seeded setting. The final row reports the RMS EO gap. Setting=Seeded contaminated.}
\label{tab:joint-eo-profession-iterative-setting}
\end{table}

\newpage

\begin{table}
\centering
\begin{tabular}{lrrrrrrr}
\toprule
\textbf{Profession} & \textbf{PT} & \textbf{Iter0 (Human)} & \textbf{Iter1} & \textbf{Iter2} & \textbf{Iter3} & \textbf{Iter4} & \textbf{Iter5} \\
\midrule
Accountant & 2.21 & -0.43 & -0.20 & 0.59 & -0.20 & -0.06 & 0.09 \\
Architect & 2.94 & 4.47 & 4.96 & 2.69 & 0.71 & 0.00 & -0.13 \\
Attorney & -0.06 & 20.50 & 40.55 & 43.81 & 7.34 & 4.88 & 0.20 \\
Chiropractor & 3.50 & -2.64 & 1.30 & -12.71 & 0.07 & 2.61 & -9.41 \\
Comedian & 0.00 & -0.36 & 0.00 & 0.00 & 0.00 & 0.00 & 0.00 \\
Composer & -0.14 & -0.48 & 1.40 & 0.88 & 0.88 & -0.14 & -0.95 \\
Dentist & -4.82 & -2.34 & -0.49 & -0.52 & 0.10 & -1.55 & -4.92 \\
Dietitian & 9.03 & 0.85 & 0.19 & -1.36 & -7.97 & -12.89 & -12.09 \\
Dj & -2.37 & -0.63 & -0.63 & -0.63 & -0.63 & 0.00 & 0.00 \\
Filmmaker & 1.77 & -2.64 & -8.96 & -5.89 & -10.48 & -14.69 & -16.00 \\
Interior Designer & -14.97 & -30.07 & -38.23 & -25.44 & -18.78 & -12.11 & -7.35 \\
Journalist & -0.18 & -0.36 & -0.17 & 0.06 & -0.35 & -0.55 & 2.48 \\
Model & 26.71 & 6.58 & 24.25 & 32.05 & 24.14 & 19.70 & 22.20 \\
Nurse & -3.85 & -8.75 & -18.80 & -30.30 & -32.03 & -29.66 & -28.72 \\
Painter & -0.73 & -2.22 & -1.46 & -1.32 & -0.66 & -1.95 & -2.31 \\
Paralegal & -5.70 & -21.21 & -23.53 & -23.17 & -20.86 & -13.90 & 7.84 \\
Pastor & 30.29 & 43.75 & 41.34 & 41.89 & 40.20 & 38.75 & 36.10 \\
Personal Trainer & 7.61 & 4.94 & 12.35 & 30.86 & 39.09 & 31.48 & 31.87 \\
Photographer & -0.19 & 0.00 & 0.00 & -0.16 & -0.16 & -0.16 & -0.16 \\
Physician & 8.30 & 14.79 & 7.29 & 1.21 & 1.82 & 2.84 & 5.27 \\
Poet & 2.53 & 0.95 & 1.32 & 0.47 & -1.04 & -1.98 & -6.05 \\
Professor & 23.18 & 21.72 & 23.71 & 31.16 & 21.86 & 8.00 & 4.22 \\
Psychologist & -5.89 & -3.56 & -6.20 & -11.04 & -8.54 & -11.53 & -6.40 \\
Rapper & 10.87 & 15.26 & 15.90 & 27.03 & 21.33 & 17.54 & 2.98 \\
Software Engineer & -7.79 & -5.04 & 0.70 & 4.43 & 0.22 & 2.17 & -1.68 \\
Surgeon & 3.59 & 4.42 & 3.00 & 6.60 & 2.40 & 3.59 & 0.42 \\
Teacher & -4.19 & -0.75 & -0.59 & -7.63 & -6.89 & -19.01 & -25.44 \\
Yoga Teacher & 4.87 & -1.61 & -1.11 & -3.39 & -3.96 & -3.96 & -0.84 \\
\midrule
RMS Gap & 10.32 & 13.18 & 16.36 & 18.82 & 15.64 & 13.93 & 13.45 \\
% Balanced RMS Gap   & 9.49 & 10.85 & 14.84 & 17.41 & 11.84 & 10.98 & 11.53 \\
\bottomrule
\end{tabular}
\caption{Equal opportunity gaps by profession across model iterations, in percentage points.  Setting=Few-shot contaminated.}
\label{tab:joint-eo-profession-fewshot}
\end{table}

\begin{table}
\centering
\begin{tabular}{lrrrrrrr}
\toprule
\textbf{Profession} & \textbf{PT} & \textbf{Iter0 (Human)} & \textbf{Iter1} & \textbf{Iter2} & \textbf{Iter3} & \textbf{Iter4} & \textbf{Iter5} \\
\midrule
Accountant & 2.21 & -0.43 & 1.02 & 0.92 & 1.82 & 1.25 & 0.53 \\
Architect & 2.94 & 4.47 & 4.86 & 5.96 & 2.98 & 5.65 & 4.70 \\
Attorney & -0.06 & 20.50 & 20.96 & 42.29 & 20.39 & 15.48 & 12.38 \\
Chiropractor & 3.50 & -2.64 & 0.24 & -11.93 & -10.10 & -9.64 & -17.30 \\
Comedian & 0.00 & -0.36 & 0.00 & -0.36 & -0.36 & 0.00 & 0.00 \\
Composer & -0.14 & -0.48 & -1.32 & -1.00 & 0.74 & 1.97 & 1.09 \\
Dentist & -4.82 & -2.34 & -5.59 & -5.08 & -10.36 & -9.12 & -10.02 \\
Dietitian & 9.03 & 0.85 & 0.19 & 0.19 & 2.83 & -0.24 & -1.57 \\
Dj & -2.37 & -0.63 & -0.63 & -0.63 & -0.63 & -1.26 & -1.26 \\
Filmmaker & 1.77 & -2.64 & 0.92 & 7.16 & 5.47 & 2.17 & -1.42 \\
Interior Designer & -14.97 & -30.07 & -31.56 & -48.44 & -28.84 & -36.73 & -35.92 \\
Journalist & -0.18 & -0.36 & 0.06 & -0.15 & -0.74 & -0.35 & 0.48 \\
Model & 26.71 & 6.58 & 45.98 & -11.91 & -1.08 & 10.48 & 34.75 \\
Nurse & -3.85 & -8.75 & -27.18 & -14.84 & -8.10 & -49.38 & -11.71 \\
Painter & -0.73 & -2.22 & -2.46 & -3.38 & -0.41 & -0.90 & -3.87 \\
Paralegal & -5.70 & -21.21 & -16.58 & -17.47 & -24.78 & -14.26 & -16.22 \\
Pastor & 30.29 & 43.75 & 43.64 & 40.72 & 38.51 & 40.50 & 46.69 \\
Personal Trainer & 7.61 & 4.94 & 3.70 & 3.70 & 2.47 & 3.70 & 2.47 \\
Photographer & -0.19 & 0.00 & -0.16 & 0.00 & -0.47 & 0.00 & 0.00 \\
Physician & 8.30 & 14.79 & 8.10 & 18.02 & 15.19 & 10.73 & 18.02 \\
Poet & 2.53 & 0.95 & -0.49 & 0.79 & -0.31 & 0.49 & 0.33 \\
Professor & 23.18 & 21.72 & 41.86 & 25.13 & 38.61 & 48.44 & 49.19 \\
Psychologist & -5.89 & -3.56 & -6.52 & -16.11 & -3.78 & -9.99 & -13.17 \\
Rapper & 10.87 & 15.26 & 10.83 & -0.22 & 10.65 & 20.40 & 15.15 \\
Software Engineer & -7.79 & -5.04 & -5.78 & -18.70 & -6.21 & -8.47 & -1.04 \\
Surgeon & 3.59 & 4.42 & 3.94 & 8.09 & 3.03 & 3.74 & 3.29 \\
Teacher & -4.19 & -0.75 & -3.10 & -0.75 & -3.02 & -2.85 & -4.69 \\
Yoga Teacher & 4.87 & -1.61 & 0.64 & -10.14 & -4.43 & -11.55 & -11.25 \\
\midrule
RMS Gap & 10.32 & 13.18 & 17.51 & 17.40 & 14.15 & 18.34 & 17.92 \\
\bottomrule
\end{tabular}
\caption{Equal opportunity gaps by profession across model iterations, in percentage points. Setting= Seeded not contaminated.}
\label{tab:joint-eo-profession-iters-not-contaminated}
\end{table}

\begin{table}
\centering
\begin{tabular}{lrrrrrrr}
\toprule
\textbf{Profession} & \textbf{PT} & \textbf{Iter0 (Human)} & \textbf{Iter1} & \textbf{Iter2} & \textbf{Iter3} & \textbf{Iter4} & \textbf{Iter5} \\
\midrule

Accountant & 2.21 & -0.43 & -0.20 & -0.20 & 0.98 & -0.08 & 0.31 \\
Architect & 2.94 & 4.47 & 4.96 & 7.00 & 5.70 & 9.06 & 8.58 \\
Attorney & -0.06 & 20.50 & 40.81 & 36.45 & 46.17 & 45.10 & 49.72 \\
Chiropractor & 3.50 & -2.64 & 1.71 & 2.44 & -3.87 & 2.28 & 0.97 \\
Comedian & 0.00 & -0.36 & 0.00 & 0.00 & 0.00 & 0.00 & 0.00 \\
Composer & -0.14 & -0.48 & 1.40 & 1.57 & 1.40 & 1.40 & 1.40 \\
Dentist & -4.82 & -2.34 & -0.49 & -4.35 & -0.77 & -1.73 & -0.28 \\
Dietitian & 9.03 & 0.85 & 0.19 & 0.85 & 0.41 & -0.90 & -0.90 \\
Dj & -2.37 & -0.63 & -0.63 & -1.26 & -0.63 & -0.63 & -0.63 \\
Filmmaker & 1.77 & -2.64 & -8.96 & -3.73 & -4.56 & -3.82 & -4.42 \\
Interior Designer & -14.97 & -30.07 & -38.23 & -30.61 & -37.01 & -28.16 & -31.84 \\
Journalist & -0.18 & -0.36 & -0.17 & -0.95 & -0.95 & -1.36 & -0.77 \\
Model & 26.71 & 6.58 & 24.25 & 8.35 & 33.52 & -26.13 & -21.71 \\
Nurse & -3.85 & -8.75 & -18.80 & -6.27 & -10.91 & -5.96 & -10.39 \\
Painter & -0.73 & -2.22 & -1.46 & -2.74 & -2.77 & -1.96 & -2.34 \\
Paralegal & -5.70 & -21.21 & -23.53 & -25.85 & -30.30 & -29.23 & -34.40 \\
Pastor & 30.29 & 43.75 & 41.34 & 42.46 & 43.00 & 40.37 & 52.61 \\
Personal Trainer & 7.61 & 4.94 & 12.35 & 9.88 & 7.41 & 6.17 & 4.94 \\
Photographer & -0.19 & 0.00 & 0.00 & 0.12 & 0.12 & 0.00 & -0.16 \\
Physician & 8.30 & 14.79 & 7.29 & 29.16 & 7.90 & 51.65 & 16.60 \\
Poet & 2.53 & 0.95 & 1.32 & 0.46 & 2.09 & 0.74 & -0.12 \\
Professor & 23.18 & 21.72 & 23.71 & 36.38 & 35.12 & 34.02 & 33.90 \\
Psychologist & -5.89 & -3.56 & -6.46 & -2.30 & -6.25 & -1.62 & -5.92 \\
Rapper & 10.87 & 15.26 & 15.90 & 17.24 & 13.63 & 27.74 & 19.32 \\
Software Engineer & -7.79 & -5.04 & 0.56 & 3.10 & 1.99 & 2.87 & 3.38 \\
Surgeon & 3.59 & 4.42 & 2.88 & 4.09 & 4.86 & 5.35 & 2.25 \\
Teacher & -4.19 & -0.75 & -0.59 & -3.44 & -4.86 & -1.93 & -1.68 \\
Yoga Teacher & 4.87 & -1.61 & -1.11 & -2.25 & -1.68 & -0.54 & 2.32 \\
\midrule
RMS Gap & 10.32 & 13.18 & 16.39 & 16.44 & 18.15 & 19.66 & 18.90 \\
% Balanced RMS Gap   & 9.49 & 10.85 & 14.89 & 16.53 & 17.62 & 21.08 & 18.16 \\
\bottomrule
\end{tabular}
\caption{Equal opportunity gaps by profession across model iterations, in percentage points. The final row reports the RMS EO gap. Setting= fewshot not contaminated.}
\label{tab:joint-eo-profession-iters-fewshot-not-contaminated}
\end{table}

\begin{table*}[t]
\centering
\setlength{\tabcolsep}{5pt}
\renewcommand{\arraystretch}{1.08}

\begin{tabular}{lrrrrrrr}
\toprule
\textbf{Profession} & \textbf{PT} & \textbf{Iter0} & \textbf{Iter1} & \textbf{Iter2} & \textbf{Iter3} & \textbf{Iter4} & \textbf{Iter5} \\
\midrule
Accountant & -0.0422 & -0.1401 & -0.1468 & -0.1528 & -0.1621 & -0.1589 & -0.1682 \\
Architect & +0.0330 & +0.0184 & +0.0131 & +0.0026 & +0.0074 & +0.0115 & -0.0034 \\
Attorney & -0.0258 & -0.0186 & -0.0150 & -0.0226 & -0.0198 & -0.0203 & -0.0260 \\
Chiropractor & +0.1521 & +0.4210 & +0.4378 & +0.4422 & +0.4527 & +0.4583 & +0.4647 \\
Comedian & +0.0566 & +0.1010 & +0.1235 & +0.1654 & +0.1598 & +0.1700 & +0.1716 \\
Composer & +0.0298 & +0.0195 & +0.0220 & +0.0167 & +0.0101 & +0.0073 & +0.0195 \\
Dentist & +0.0850 & -0.0379 & -0.0319 & -0.0325 & -0.0310 & -0.0391 & -0.0367 \\
Dietitian & -0.1186 & -0.5006 & -0.5113 & -0.5217 & -0.5384 & -0.5442 & -0.5688 \\
Dj & -0.1345 & -0.0777 & -0.0628 & -0.0510 & -0.0323 & -0.0278 & -0.0247 \\
Filmmaker & -0.0440 & -0.0533 & -0.0604 & -0.0679 & -0.0738 & -0.0790 & -0.0772 \\
Interior Designer & -0.1170 & -0.1332 & -0.1352 & -0.1480 & -0.1463 & -0.1478 & -0.1303 \\
Journalist & -0.0514 & -0.0828 & -0.0880 & -0.1060 & -0.1180 & -0.1158 & -0.1139 \\
Model & -0.4672 & -2.0250 & -2.1131 & -2.1910 & -2.2485 & -2.3056 & -2.3563 \\
Nurse & -0.0933 & +0.0788 & +0.0843 & +0.0873 & +0.0862 & +0.0900 & +0.0876 \\
Painter & -0.0558 & -0.0945 & -0.1057 & -0.1027 & -0.1066 & -0.1056 & -0.1138 \\
Paralegal & +0.0373 & -0.1244 & -0.1395 & -0.1696 & -0.1757 & -0.1949 & -0.2058 \\
Pastor & +0.1001 & +0.0926 & +0.0817 & +0.0892 & +0.0902 & +0.0800 & +0.0948 \\
Personal Trainer & -0.0963 & -0.0908 & -0.0975 & -0.1010 & -0.1093 & -0.1104 & -0.1065 \\
Photographer & -0.0543 & -0.0483 & -0.0535 & -0.0547 & -0.0583 & -0.0623 & -0.0643 \\
Physician & -0.0110 & +0.2089 & +0.2211 & +0.2168 & +0.2293 & +0.2321 & +0.2374 \\
Poet & -0.0710 & -0.1019 & -0.1118 & -0.1153 & -0.1260 & -0.1248 & -0.1275 \\
Professor & +0.0729 & +0.0347 & +0.0376 & +0.0392 & +0.0419 & +0.0431 & +0.0494 \\
Psychologist & -0.0961 & +0.0904 & +0.0972 & +0.0975 & +0.1001 & +0.1031 & +0.1090 \\
Rapper & -0.1339 & -0.0987 & -0.0901 & -0.0801 & -0.0745 & -0.0727 & -0.0433 \\
Software Engineer & +0.0923 & +0.0849 & +0.0781 & +0.0717 & +0.0774 & +0.0746 & +0.0820 \\
Surgeon & +0.0166 & +0.1454 & +0.1645 & +0.1752 & +0.1813 & +0.1953 & +0.2006 \\
Teacher & -0.1502 & -0.1736 & -0.1784 & -0.1905 & -0.1993 & -0.2072 & -0.2100 \\
Yoga Teacher & +0.0130 & -0.0889 & -0.1059 & -0.1008 & -0.1041 & -0.1144 & -0.1115 \\
\bottomrule
\end{tabular}

\caption{
Signed gender-conditioned NLL gap by profession across iterations under the contaminated few-shot regime. Values report $\Delta\mathrm{NLL}=\mathrm{NLL}_{female}-\mathrm{NLL}_{male}$. Positive values indicate female biographies are less likely under the model; negative values indicate male biographies are less likely.
Setting=fewshot, contaminated.
}
\label{tab:all_professions_nll_gap_fewshot_contam}
\end{table*}

\begin{table*}[t]
\centering
\setlength{\tabcolsep}{5pt}
\renewcommand{\arraystretch}{1.08}

\begin{tabular}{lrrrrrrr}
\toprule
\textbf{Profession} & \textbf{PT} & \textbf{Iter0} & \textbf{Iter1} & \textbf{Iter2} & \textbf{Iter3} & \textbf{Iter4} & \textbf{Iter5} \\
\midrule
Accountant & -0.0422 & -0.1401 & -0.1427 & -0.1533 & -0.1606 & -0.1663 & -0.1673 \\
Architect & +0.0330 & +0.0184 & +0.0149 & +0.0058 & +0.0021 & +0.0017 & -0.0015 \\
Attorney & -0.0258 & -0.0186 & -0.0164 & -0.0235 & -0.0226 & -0.0268 & -0.0237 \\
Chiropractor & +0.1521 & +0.4210 & +0.4416 & +0.4455 & +0.4503 & +0.4728 & +0.4857 \\
Comedian & +0.0566 & +0.1010 & +0.1214 & +0.1400 & +0.1488 & +0.1621 & +0.1870 \\
Composer & +0.0298 & +0.0195 & +0.0104 & +0.0099 & +0.0197 & +0.0265 & +0.0255 \\
Dentist & +0.0850 & -0.0379 & -0.0429 & -0.0387 & -0.0356 & -0.0347 & -0.0293 \\
Dietitian & -0.1186 & -0.5006 & -0.5065 & -0.5220 & -0.5333 & -0.5676 & -0.5704 \\
Dj & -0.1345 & -0.0777 & -0.0621 & -0.0862 & -0.0650 & -0.0762 & -0.0321 \\
Filmmaker & -0.0440 & -0.0533 & -0.0572 & -0.0609 & -0.0688 & -0.0785 & -0.0766 \\
Interior Designer & -0.1170 & -0.1332 & -0.1277 & -0.1301 & -0.1444 & -0.1303 & -0.1577 \\
Journalist & -0.0514 & -0.0828 & -0.0921 & -0.1033 & -0.1090 & -0.1182 & -0.1189 \\
Model & -0.4672 & -2.0250 & -2.1196 & -2.1933 & -2.2479 & -2.2980 & -2.3446 \\
Nurse & -0.0933 & +0.0788 & +0.0793 & +0.0880 & +0.0857 & +0.0824 & +0.0832 \\
Painter & -0.0558 & -0.0945 & -0.1037 & -0.1089 & -0.1062 & -0.1115 & -0.1196 \\
Paralegal & +0.0373 & -0.1244 & -0.1408 & -0.1599 & -0.2037 & -0.2011 & -0.2099 \\
Pastor & +0.1001 & +0.0926 & +0.0873 & +0.0937 & +0.0683 & +0.0772 & +0.0692 \\
Personal Trainer & -0.0963 & -0.0908 & -0.0976 & -0.1045 & -0.1093 & -0.0960 & -0.1059 \\
Photographer & -0.0543 & -0.0483 & -0.0529 & -0.0571 & -0.0596 & -0.0571 & -0.0620 \\
Physician & -0.0110 & +0.2089 & +0.2162 & +0.2251 & +0.2265 & +0.2325 & +0.2373 \\
Poet & -0.0710 & -0.1019 & -0.1114 & -0.1145 & -0.1218 & -0.1337 & -0.1349 \\
Professor & +0.0729 & +0.0347 & +0.0340 & +0.0378 & +0.0343 & +0.0438 & +0.0384 \\
Psychologist & -0.0961 & +0.0904 & +0.0946 & +0.1010 & +0.1012 & +0.1004 & +0.1037 \\
Rapper & -0.1339 & -0.0987 & -0.1015 & -0.0940 & -0.0204 & -0.0372 & -0.0183 \\
Software Engineer & +0.0923 & +0.0849 & +0.0810 & +0.0861 & +0.0824 & +0.0714 & +0.0857 \\
Surgeon & +0.0166 & +0.1454 & +0.1603 & +0.1801 & +0.1903 & +0.1917 & +0.1970 \\
Teacher & -0.1502 & -0.1736 & -0.1787 & -0.1904 & -0.2059 & -0.2061 & -0.2126 \\
Yoga Teacher & +0.0130 & -0.0889 & -0.1142 & -0.1266 & -0.1265 & -0.1038 & -0.1036 \\
\bottomrule
\end{tabular}

\caption{
Signed gender-conditioned NLL gap by profession across model iterations. Values report $\Delta\mathrm{NLL}=\mathrm{NLL}_{female}-\mathrm{NLL}_{male}$. Positive values indicate female biographies are less likely under the model; negative values indicate male biographies are less likely. Setting=seeded, contaminated.
}
\label{tab:nll_gap_all_professions_seeded_contaminated}
\end{table*}

\begin{table*}[t]
\centering
\setlength{\tabcolsep}{5pt}
\renewcommand{\arraystretch}{1.08}

\begin{tabular}{lrrrrr}
\toprule
\textbf{Profession} & \textbf{Iter1} & \textbf{Iter2} & \textbf{Iter3} & \textbf{Iter4} & \textbf{Iter5} \\
\midrule
Accountant & -0.1468 & -0.1446 & -0.1446 & -0.1428 & -0.1429 \\
Architect & +0.0131 & +0.0119 & +0.0084 & +0.0127 & +0.0104 \\
Attorney & -0.0150 & -0.0214 & -0.0211 & -0.0180 & -0.0200 \\
Chiropractor & +0.4378 & +0.4393 & +0.4346 & +0.4430 & +0.4406 \\
Comedian & +0.1235 & +0.1294 & +0.1407 & +0.1330 & +0.1325 \\
Composer & +0.0220 & +0.0237 & +0.0224 & +0.0228 & +0.0191 \\
Dentist & -0.0319 & -0.0360 & -0.0368 & -0.0333 & -0.0338 \\
Dietitian & -0.5113 & -0.5148 & -0.5132 & -0.5043 & -0.5067 \\
Dj & -0.0628 & -0.0636 & -0.0561 & -0.0676 & -0.0627 \\
Filmmaker & -0.0604 & -0.0588 & -0.0603 & -0.0580 & -0.0585 \\
Interior Designer & -0.1352 & -0.1246 & -0.1357 & -0.1261 & -0.1272 \\
Journalist & -0.0880 & -0.0938 & -0.0951 & -0.0936 & -0.0925 \\
Model & -2.1131 & -2.1171 & -2.1181 & -2.1146 & -2.1178 \\
Nurse & +0.0843 & +0.0737 & +0.0834 & +0.0777 & +0.0812 \\
Painter & -0.1057 & -0.0995 & -0.1008 & -0.1011 & -0.1002 \\
Paralegal & -0.1395 & -0.1391 & -0.1515 & -0.1378 & -0.1409 \\
Pastor & +0.0817 & +0.0755 & +0.0796 & +0.0817 & +0.0869 \\
Personal Trainer & -0.0975 & -0.0999 & -0.1010 & -0.0940 & -0.1036 \\
Photographer & -0.0535 & -0.0542 & -0.0500 & -0.0568 & -0.0538 \\
Physician & +0.2211 & +0.2216 & +0.2172 & +0.2228 & +0.2203 \\
Poet & -0.1118 & -0.1052 & -0.1086 & -0.1070 & -0.1113 \\
Professor & +0.0376 & +0.0383 & +0.0354 & +0.0336 & +0.0358 \\
Psychologist & +0.0972 & +0.0967 & +0.0937 & +0.0972 & +0.0961 \\
Rapper & -0.0901 & -0.0993 & -0.0897 & -0.0939 & -0.0707 \\
Software Engineer & +0.0781 & +0.0827 & +0.0846 & +0.0819 & +0.0821 \\
Surgeon & +0.1645 & +0.1592 & +0.1572 & +0.1609 & +0.1583 \\
Teacher & -0.1784 & -0.1822 & -0.1788 & -0.1816 & -0.1849 \\
Yoga Teacher & -0.1059 & -0.0854 & -0.0989 & -0.1046 & -0.0939 \\
\bottomrule
\end{tabular}

\caption{
Signed gender-conditioned NLL gap by profession across iterations under the \textbf{iterative few-shot} regime.
Values report $\Delta\mathrm{NLL}=\mathrm{NLL}_{female}-\mathrm{NLL}_{male}$.
Positive values indicate female biographies are less likely under the model; negative values indicate male biographies are less likely. Setting=few-shot, iterative.
}
\label{tab:all_professions_nll_gap_fewshot_iterative}
\end{table*}

\begin{table*}[t]
\centering
\setlength{\tabcolsep}{5pt}
\renewcommand{\arraystretch}{1.08}

\begin{tabular}{lrrrrr}
\toprule
\textbf{Profession} & \textbf{Iter1} & \textbf{Iter2} & \textbf{Iter3} & \textbf{Iter4} & \textbf{Iter5} \\
\midrule
Accountant & -0.1392 & -0.1382 & -0.1414 & -0.1352 & -0.1399 \\
Architect & +0.0309 & +0.0195 & +0.0273 & +0.0222 & +0.0166 \\
Attorney & -0.0160 & -0.0190 & -0.0205 & -0.0154 & -0.0177 \\
Chiropractor & +0.4268 & +0.4213 & +0.4263 & +0.4207 & +0.4213 \\
Comedian & +0.0985 & +0.0988 & +0.0942 & +0.0824 & +0.0854 \\
Composer & +0.0137 & +0.0167 & +0.0275 & +0.0155 & +0.0178 \\
Dentist & -0.0379 & -0.0369 & -0.0222 & -0.0340 & -0.0402 \\
Dietitian & -0.5055 & -0.5094 & -0.5111 & -0.4946 & -0.5046 \\
Dj & -0.0750 & -0.0646 & -0.0612 & -0.0535 & -0.0739 \\
Filmmaker & -0.0512 & -0.0544 & -0.0531 & -0.0534 & -0.0525 \\
Interior Designer & -0.1343 & -0.1212 & -0.1220 & -0.1305 & -0.1218 \\
Journalist & -0.0813 & -0.0860 & -0.0857 & -0.0857 & -0.0861 \\
Model & -2.0408 & -2.0454 & -2.0393 & -2.0426 & -2.0485 \\
Nurse & +0.0803 & +0.0806 & +0.0845 & +0.0743 & +0.0796 \\
Painter & -0.0880 & -0.0941 & -0.0877 & -0.0930 & -0.0965 \\
Paralegal & -0.1158 & -0.1121 & -0.1056 & -0.1069 & -0.1092 \\
Pastor & +0.0972 & +0.0869 & +0.0873 & +0.0874 & +0.0923 \\
Personal Trainer & -0.0888 & -0.1011 & -0.0864 & -0.0935 & -0.1073 \\
Photographer & -0.0441 & -0.0438 & -0.0485 & -0.0474 & -0.0428 \\
Physician & +0.2054 & +0.2060 & +0.2094 & +0.2084 & +0.2101 \\
Poet & -0.1110 & -0.0954 & -0.1047 & -0.1075 & -0.1011 \\
Professor & +0.0341 & +0.0369 & +0.0326 & +0.0314 & +0.0295 \\
Psychologist & +0.0920 & +0.0965 & +0.1000 & +0.0916 & +0.0958 \\
Rapper & -0.1138 & -0.0834 & -0.0855 & -0.1023 & -0.1189 \\
Software Engineer & +0.0864 & +0.0851 & +0.0879 & +0.0810 & +0.0832 \\
Surgeon & +0.1461 & +0.1449 & +0.1509 & +0.1464 & +0.1468 \\
Teacher & -0.1724 & -0.1705 & -0.1737 & -0.1713 & -0.1728 \\
Yoga Teacher & -0.0953 & -0.1087 & -0.0953 & -0.0944 & -0.0839 \\
\bottomrule
\end{tabular}

\caption{
Signed gender-conditioned NLL gap by profession across iterations under the \textbf{seeded iterative} regime.
Values report $\Delta\mathrm{NLL}=\mathrm{NLL}_{female}-\mathrm{NLL}_{male}$.
Positive values indicate female biographies are less likely under the model; negative values indicate male biographies are less likely. Setting=seeded, iterative.
}
\label{tab:all_professions_nll_gap_seeded_iterative}
\end{table*}

\begin{table*}[t]
\centering
\footnotesize
\setlength{\tabcolsep}{6pt}
\renewcommand{\arraystretch}{1.14}

\begin{tabular}{>{\raggedright\arraybackslash}p{0.94\textwidth}}
\toprule

\textbf{Prompt + seeded context.}
\textit{Continue the following professional biography in the same factual, third-person style:}
She received her BS degree in Nutrition and Food Science from Utah State University.
She then went on to complete her Dietetic Internship and receive her Master
\\
\midrule

\textbf{\textcolor{ANRblue}{Human.}}
\ldots{}'s Degree from Stony Brook University in New York.
She studied the Mediterranean Diet in Italy, where her love for Italian cuisine grew.
She is also a certified Pilates instructor, teaching at Studio Core Pilates.
Devrie has worked at Life Time Fitness as the Nutrition Coach and co-developed a corporate wellness program for the top insurance brokerage firm in Utah.
Devrie co-founded a successful online business named PureFit PureFood in 2014 and has re-branded it to support a more concise wellness message focused on Intuitive Eating, HAES, and restriction recovery.
She does not believe in diets, loves to cook, and tries new restaurants.
Devrie is a wife and mother of four.
Raising them as ``foodies'' rather than ``health nuts'' is one of her deepest passions.
\\
\midrule

\textbf{\textcolor{red!70!black}{Iteration 1.}}
\ldots{}'s Degree at Brigham Young University.
After graduating with a Certificate of Training in Sports and Athletic Nutrition from Northwestern Memorial Hospital, her current research interests include sports nutrition and sports injury prevention.
In addition, she has worked as an adjunct professor for several colleges in Utah.
Dr. Rittenhouse is the author of several books including \emph{The Power of Foods} and \emph{Fitness for Sport and Competition}.
Read more on Last.fm.
User-contributed text is available under the Creative Commons By-SA License; additional terms may apply.
\\
\midrule

\textbf{\textcolor{red!70!black}{Iteration 2.}}
\ldots{}'s Degree in Clinical Nutrition from Utah State University.
After moving to Colorado, she became a licensed dietitian at The Boulder Hospital for Children and continued there with an internship.
In 2014, Dr. Poythewalley went back to New York City to pursue a graduate degree in dietetics at Columbia University College of Physicians and Surgeons.
After graduating in 2017, Dr. Poythewalley completed her dietetic residency and served as an instructor of nutrition.
\\
\midrule

\textbf{\textcolor{red!70!black}{Iteration 3.}}
\ldots{}of Science degree in Nutrition and Food Science from Ohio State University.
She completed pre- and post-internship clinical dietetic training at the University of Iowa Hospital for Children and later completed a post-internship position at Cleveland Clinic as an instructor of nutrition science.
Her areas of interest include obesity prevention and treatment, hypertension and diabetes management, food allergies, nutritional counseling for children with developmental eating issues, and healthy eating trends.
\\
\midrule

\textbf{\textcolor{red!70!black}{Iteration 4.}}
\ldots{}'s degree in Clinical Nutrition from The University of Texas at Houston College of Health Sciences.
After completing the internship, she worked as a clinical dietitian for several organizations including The National Institute of Health, The American Dietetic Association and the International Society of Dietitians \& Educators.
Her current practice includes providing nutrition counseling services to clients interested in weight loss management, food allergies, diabetes prevention, and sports performance.
To learn more about Dr. Gossen's practice experience, visit \url{www.dieting.org}.
Continue reading \ldots{} Continue reading \ldots{}
\\
\midrule

\textbf{\textcolor{red!70!black}{Iteration 5.}}
\ldots{}'s Degree in Health Promotion Management from the College of Southern California.
This was followed by a year-long postdoctoral fellowship in clinical nutrition at Stanford University's School of Medicine, where she served as an associate professor of medicine.
She currently works for a private practice specializing in weight management programs for individuals seeking help with eating disorders, diabetes-related conditions, hypertension, and cardiovascular disease.
In addition to being a licensed dietitian practitioner, she also holds several fabricated-sounding degrees and credentials, while the continuation ends with generic web-page artifacts such as ``Read more,'' ``Related Links,'' and numbered links.
\\

\bottomrule
\end{tabular}

\caption{
An example of model continuations over recursive training iterations.
A single biography cannot fully capture distribution-level changes, but it can reveal recurring qualitative patterns.
In this example, the human continuation gives specific details about a dietitian and wellness professional.
After several iterations of training on synthetic text, the model increasingly changes institutions and credentials, shifts toward a more generic clinical framing, and produces less faithful biographical details and web-page artifacts.
}
\label{tab:example-biography-recursive}
\end{table*}

\begin{table*}[t]
\centering 
\small
\setlength{\tabcolsep}{4pt}
\renewcommand{\arraystretch}{1.08}

\begin{tabular}{lccccc}
\toprule
\multirow{2}{*}{\textbf{Iteration}} 
& \multirow{2}{*}{\textbf{PPL}$\downarrow$}
& \multirow{2}{*}{\textbf{MMLU}$\uparrow$}
& \multicolumn{3}{c}{\textbf{Bias-in-Bios}} \\
\cmidrule(lr){4-6}
& 
& 
& \textbf{NLL-GAP}$\downarrow$  
& \textbf{Acc.}$\uparrow$
& \textbf{EO GAP}$\downarrow$ \\
\midrule
Base & -     & 44.38 & -0.038 & 83.37 & 10.32 \\
1    & 16.07 & 42.14 & -0.093 & 80.05 & 13.18 \\
2    & 15.01 & 40.31 & -0.096 & 81.10 & 12.57 \\
3    & 14.20 & 38.19 & -0.098 & 80.11 & 11.31 \\
4    & 12.26 & 36.29 & -0.101 & 79.14     & 12.44     \\
5    & 12.01 & 34.57 & -0.103 & 74.28     & 12.57     \\
\bottomrule
\end{tabular}

\caption{
Perplexity, MMLU, and Bias-in-Bios results for models trained on the \textbf{human-written} corpus.
PPL is measured on the same human-written corpus used for training.
Bias-in-Bios includes the signed gender-conditioned NLL gap, occupation-prediction accuracy, and equal-opportunity gap across professions.
Bias-in-Bios accuracy fluctuates until iteration 4, while the EO gap stays roughly between 11 and 13.
This differs from the iterative and recursive synthetic-data regimes, where the EO gap generally increases across iterations before the MMLU drop becomes significant.
For seeded generation, the EO gap reaches about 14-18 under the iterative setting and 15-20 under the recursive setting.
For few-shot generation, it reaches about 16-19 under an iterative setting and 16-18 under a recursive setting.
}
\label{tab:human-only-multi-benchmark-results}
\end{table*}

\end{document}